\documentclass[10pt,journal,final,twocolumn]{IEEEtran}

\usepackage{graphicx,times,amsmath,amssymb,cite,cases,mathrsfs,booktabs,multirow,psfrag,subfigure,url,stfloats,threeparttable,epstopdf,pdfpages}
\usepackage{algorithm,algorithmic,amsthm,ulem,latexsym,bm}
\usepackage{dsfont}
\usepackage[colorlinks,linkcolor=red,anchorcolor=blue,citecolor=red]{hyperref}

\begin{document}

\title{Multi-Source Unsupervised Domain Adaptation via Pseudo Target Domain}
\author{Chuan-Xian Ren, Yong-Hui Liu, Xi-Wen Zhang, Ke-Kun Huang 
\thanks{This work is supported in part by the National Natural Science Foundation of China under Grant 61976229, Grant 61906046, Grant 61976104 and Grant 12026601, in part by the Open Research Projects of Zhejiang Lab (Grant 2021KH0AB08), and in part by Guangdong-Hong Kong-Macau Applied Math Center (Grant 2020B1515310011). (Corresponding author: Chuan-Xian Ren)}
\thanks{C.-X. Ren, Y.-H. Liu and X.-W. Zhang are with the School of Mathematics, Sun Yat-sen University, Guangzhou, 510275, China (email: rchuanx@mail.sysu.edu.cn).} \thanks{K.-K. Huang is with the School of Mathematics, JiaYing University, Meizhou, 514015, China.}}

\date{}
\IEEEcompsoctitleabstractindextext{%

\begin{abstract}
Multi-source domain adaptation (MDA) aims to transfer knowledge from multiple source domains to an unlabeled target domain. MDA is a challenging task due to the severe domain shift, which not only exists between target and source but also exists among diverse sources. Prior studies on MDA either estimate a mixed distribution of source domains or combine multiple single-source models, but few of them delve into the relevant information among diverse source domains. For this reason, we propose a novel MDA approach, termed Pseudo Target for MDA (PTMDA). Specifically, PTMDA maps each group of source and target domains into a group-specific subspace using adversarial learning with a metric constraint, and constructs a series of pseudo target domains correspondingly. Then we align the remainder source domains with the pseudo target domain in the subspace efficiently, which allows to exploit additional structured source information through the training on pseudo target domain and improves the performance on the real target domain. Besides, to improve the transferability of deep neural networks (DNNs), we replace the traditional batch normalization layer with an effective matching normalization layer, which enforces alignments in latent layers of DNNs and thus gains further promotion. We give theoretical analysis showing that PTMDA as a whole can reduce the target error bound and leads to a better approximation of the target risk in MDA settings. Extensive experiments demonstrate PTMDA's effectiveness on MDA tasks, as it outperforms state-of-the-art methods in most experimental settings.
\end{abstract}

}

\maketitle \IEEEdisplaynotcompsoctitleabstractindextext \IEEEpeerreviewmaketitle

\section{Introduction}\label{section1}

\IEEEPARstart{D}{eep} neural networks (DNNs) are powerful at extracting features from structured data \cite{resnet2016,zhanglei2020Class} such as image, speech, and video, and it significantly outperforms traditional machine learning algorithms in image processing and classification. However, these remarkable gains often rely on the availability of large amounts of labeled training samples~\cite{Yang2020tip}, which limits the utility of DNNs in situations where sample labeling is prohibitively expensive. One natural idea to address the issue is to generalize the model from a labeled dataset to the unlabeled dataset\cite{ren2014tipTransfer,Lu2016tip}. However, if the distribution of the labeled dataset differs from that of the unlabeled dataset, the model cannot be generalized well. Unsupervised domain adaptation (UDA) \cite{Yaroslav2016JMLR,Saito2018CVPRmcd,ren2020TCYB1} attempts to address this domain shift issue, i.e. the fact that there is a distribution bias between the training and test datasets in many practical applications.

Great efforts \cite{Saito2018CVPRmcd,Kui2019CVPR,ren2019TCYB1,Long2019PAMI1,mdd2020pami1} have been devoted to the UDA literatures. While existing works mainly focus on tasks with single source domain, it is common in practice that labeled samples are obtained from multiple sources with diverse distributions. The task with such multiple source domains is known as multi-source unsupervised domain adaptation (MDA)\cite{msda2012video1,zhao2019NIPS1seg}. For instance, we want to predict the category of some photos, which can be taken as the target domain. We can search from the Internet for labeled images such as paintings, cliparts, or sketches. Those images can be taken as multiple sources, which exhibit significant difference in texture or visual style. A straightforward strategy is to combine these source domains into a single domain, but it leads to a sub-optimal model since gaps among multiple source domains are omitted. This practical issue motivates the study on MDA, which aims at learning a prediction model from multiple sources and generalizing it to a different yet related target domain.

\begin{figure}[tb]
\centering\includegraphics[scale=0.7]{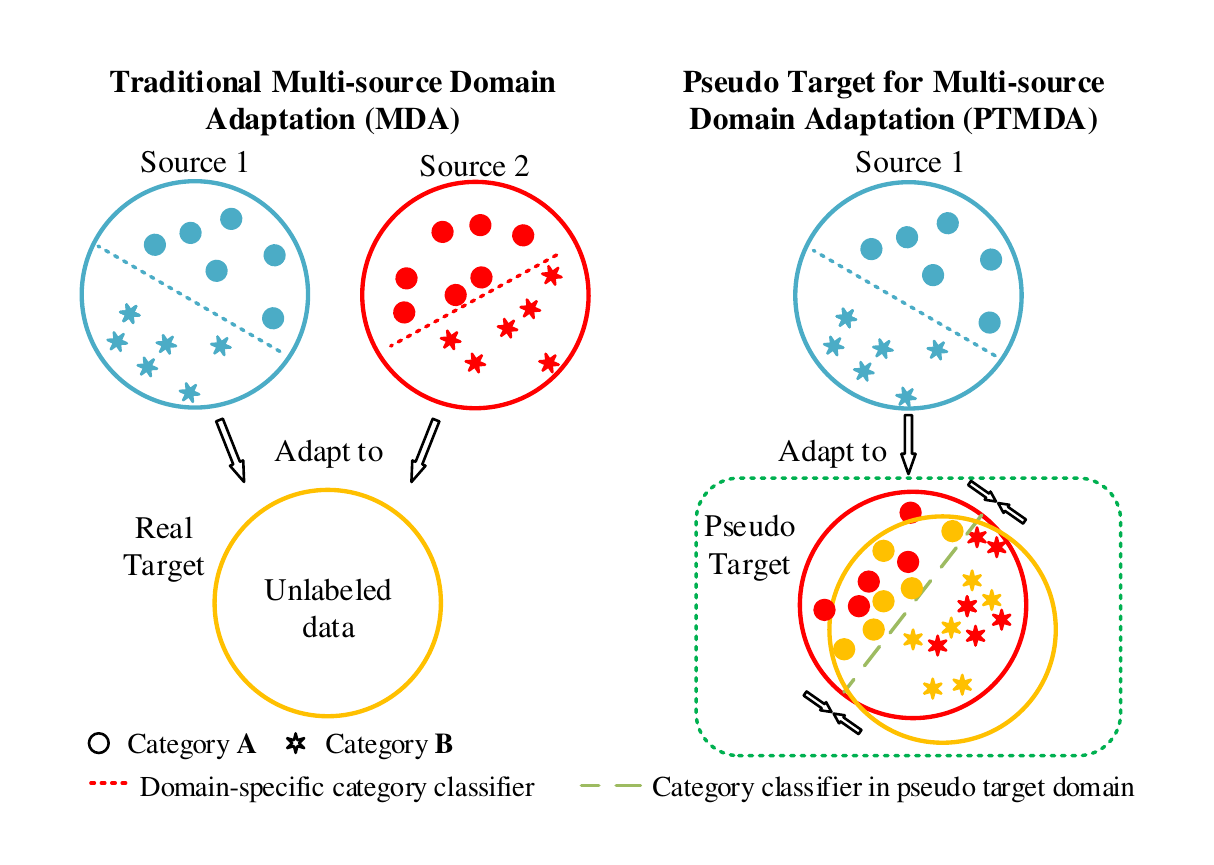}
\caption{Difference between PTMDA and traditional MDA methods. Traditional MDA methods (left side) separately align each source domain with the real target domain. PTMDA (right side) first reduces domain shift between the real target domain and source domain 2 in a group-specific subspace, then combine them to construct a \textit{pseudo target domain}, which will be aligned with source domain 1. In this way,
structured information from source domains can be rendered into target domain to facilitate discriminative feature extraction, and thus benefits generalization on target domain. Better viewed in color.}\label{fig1}
\end{figure}

Recently, based on the distribution weighting rule of Mansour et al.\cite{Mansour2009NIPS1}, mixed distributions{\cite{Xu_2018_CVPR,zhu2019MSDA,DistillMSDA2020AAAI}} from multiple sources are designed and theoretical results{\cite{Zhao2018NIPS,Hoffman2018NIPS,Ievgen2019On}} are derived. Therefore, combination-based MDA methods have gained popularity. These methods first train source-specific models corresponding to each pair of source and target domains by minimizing the distribution discrepancy~\cite{Hoffman2012LtnDm,Peng_2019_ICCV,zhu2019MSDA}, or by learning domain-invariant features via adversarially confusing a domain discriminator~\cite{Xu_2018_CVPR,DistillMSDA2020AAAI,chen2020Multiple}. Then, a weight vector is derived to combine the predictions for the target samples. While previous methods have already achieved competitive results on MDA, few of them delve into the relevance among diverse source domains with abundant discriminative information. Moreover, it is still important to improve the transferability of the latent layers of DNNs. Although related works~\cite{BoostMda2018,DSBN2019,MultiDial2020} using variants of the Batch Normalization (BN) layers \cite{2015Batch} have been raised, these methods achieve only limited effectiveness.

In this paper, we propose a new concept of \textit{Pseudo Target Domain} to deal with MDA problems, and thus call our method PTMDA for short. The main difference between PTMDA and traditional MDA methods is shown in Fig.~\ref{fig1}. Traditional methods (left) focus on aligning each source domain with the real target domain. Notice that while the aligning process mapping target domain and a source domain into a shared space, distribution gap between domains is reduced and the two domains are drew closer in the mapping space. It motivates us to combine the two mapped domains after the alignment (right). The pseudo target domain, which incorporates labeled samples into the unlabeled target, can be served as a new informative target to the other source domains. With the construction of pseudo target domain, PTMDA not only reduces domain shift between source and target domains, and, more importantly, it is capable to efficiently utilize discriminative information from source domain to promote downstream alignments with other source domains. Besides, we design a simple yet effective matching normalization (MN) layer to further improve the generalization ability of DNNs. We propose to update the affine parameters with gradient information only from the target domain in the training stage. This special layer aligns the distributions of source domain and target domain in the latent layers of the feature extraction network.

Our main contributions are summarized as follows.
\begin{enumerate}
\item We construct a series of pseudo target domains by performing the adversarial training strategy on the real target domain and each source domain. This approach conveys useful information from each source domain to the target domain, and thus is helpful to the alignment of the pseudo target domain with the remainder source domains. The method takes advantage of multiple source knowledge in MDA tasks and promotes the generalization performance on the real target domain.
\item We introduce the MN layer, which aligns the distributions of different domains in the latent network layers to improve the transfer ability of DNNs. The proposed MN layer is generic and can be embedded into many deep domain adaptation methods.
\item PTMDA achieves state-of-the-art classification performance for MDA tasks on five benchmark datasets. In particular, the average accuracy of MDA tasks increases by 1.2\% on Office-Caltech10 and increases by 0.8\% on ImageCLEF-DA.
\end{enumerate}

The remainder of this paper is organized as follows. In Section~\ref{section2}, we briefly review some related works. Details of the PTMDA method are given in Section~\ref{section3}. Section~\ref{sectionT} presents theoretical analysis for the effectiveness of PTMDA. Section \ref{section4} provides comprehensive experimental results for validating the effectiveness of PTMDA. Section \ref{section5} concludes the paper.

\section{Related Work}\label{section2}

In this section, we briefly review some recently proposed and related works of UDA and MDA.

\subsection{Unsupervised Domain Adaptation}

Most recent UDA methods are motivated by the results of Ben-David et al.~\cite{Ben2006NIPS}, which exploited the $\mathcal{A}$-distance to estimate the discrepancy of distributions between source and target domains. Blitzer et al.~\cite{Blitzer2007NIPS} deduced a uniform convergence learning bound, which minimizes the convex combination of empirical risks among diverse domains. By virtue of these foundation works, minimizing the domain discrepancy has been extensively used for UDA. Hu et al. \cite{Lu2016tip} proposed deep transfer metric learning and constrain the local manifold to enhance the discrimination ability of representations. Saito et al. \cite{uniDA2020} proposed to cluster the neighboring target data using self-supervision information to learn discriminative features. Maximum mean discrepancy (MMD) is an important statistic used to reduce distribution shift in various methods~\cite{Long2019PAMI1}. Ren et al. \cite{ren2020TCYB1} exploited low-rank  representation to alleviate the distribution shift and stress the group compactness of features. Some other methods \cite{Li2020etd,Kui2018IJCAI} used the idea of optimal transportation to measure the domain discrepancy. Luo et al. \cite{YouWeiPami1} proposed a discriminative manifold propagation (DMP) framework to improve the model's generalization ability on the target domain. Pandey et al. \cite{TgtInd2020} proposed to find the `closest-clone', which is a source image that arbitrarily close to the test image from the target data, and train the domain-adaptive classifier using the clones.

While many methods discussed above were proposed based on distribution discrepancy, there are several other works using the technique of adversarial training. Ganin et al.\cite{Yaroslav2016JMLR} introduced domain adversarial neural network (DANN) to align the feature distributions. Tzeng et al. \cite{Tzeng2017CVPRadda} used separate feature networks for diverse domains and train the target domain adversarially. Carlucci et al.\cite{hallu2019} and Kurmi et al.\cite{lookback2019} used multi-class discriminator to improve the performance of adversarial domain adaptation. Satio et al. \cite{Saito2018CVPRmcd} proposed the maximum classifier discrepancy (MCD), which utilizes two classifiers to simulate a domain discriminator. Zhang et al. \cite{Kui2019CVPR} designed a symmetric architecture to perform domain confusion both on category-level and domain-level. Pei et al.\cite{pei2018multi} used specialized domain discriminators for each class to enable fine-grained adaptation among various domains.

The method of pseudo labeling was also employed in some works since it is believed that the construction of pseudo labels can improve the discriminative ability of the unlabeled data. Saito et al.~\cite{Asymm2017} used three asymmetric classifiers to assign robust pseudo labels to unlabeled data. Zhang et al.~\cite{Collab2021} proposed a collaborative and adversarial network (CAN) to iteratively refine the quality of pseudo labels and then extend the network with a self-paced learning scheme~\cite{Collab2018}. Deng et al.~\cite{Trip2021} used a similarity guided constraint to progressively select pseudo labels. In our work, we will use a confidence threshold to eliminate the unreliable pseudo labels.

\subsection{Multi-source Unsupervised Domain Adaptation}

Several works have been devoted to the theoretical study for MDA, Mansour et al.~\cite{Mansour2009NIPS1} provided theoretical analysis under the assumption that the target domain is a convex combination of source domains. Hoffman et al.\cite{Hoffman2018NIPS} considered an extension of the theory of Mansour to derive the mixture parameter. Zhao et al.\cite{Zhao2018NIPS} introduced new error bounds for both regression and classification tasks. Redko et al.\cite{Ievgen2019On} used a Wasserstein distance-based error function to reformulate the join hypothesis estimation of MDA task. Wen et al.\cite{AggrMSDA2020ICLM} derived a finite-sample error bound based on the theory of Mansour et al.~\cite{Mansour2009NIPS1}.

There are other works on specific algorithms. Hoffman et al.\cite{Hoffman2012LtnDm} introduced a domain transform framework, which uses a cluster approach to discover the latent domains. Xu et al.~\cite{Xu_2018_CVPR} proposed the deep cocktail network (DCTN) using multi-way domain adversarial learning. Peng et al.~\cite{Peng_2019_ICCV} defined a cross-moment divergence to enforce alignment between each pair of domains. Zhao et al.~\cite{DistillMSDA2020AAAI} designed a novel weighting strategy to combine diverse classifiers. The recently proposed MFSAN~\cite{zhu2019MSDA} first aligns the domain-specific distribution and then matches domain-specific classifiers. These works either estimate a mixed distribution~\cite{Xu_2018_CVPR,zhu2019MSDA,DistillMSDA2020AAAI} or combine multiple single-source models~\cite{Peng_2019_ICCV,Hoffman2012LtnDm,Hoffman2018NIPS,chen2020Multiple}. Carlucci et al.~\cite{hallu2019} used a Hallucinator block to remove the domain-specific style across various source domains.

In this paper, we propose a novel method to deal with MDA problems. Our PTMDA method is different from existing MDA algorithms in the sense that, PTMDA exploits structured and relevant information among source domains in addition to improve generalization on the target domain, and improves the transferability of the intermediate layers of DNNs to mitigate the domain shift among multiple domains.

\section{Learning with Pseudo Target Domain}\label{section3}

In this section, we first present general MDA problems, and then describe the construction of pseudo target domain and the MN layers. Finally, we present the PTMDA method in details.

\subsection{Problem Formulation}

Without loss of generality, we consider a $C$-class problem. Let $\mathcal{X}$ denote the input space and $\mathcal{Y}$ denote the output space, where $\mathcal{Y} = \{1,\cdots,C\}$. We define a domain $\mathcal{D}$ enclosed with a distribution $P$ and a labeling function $f: \mathcal{X} \rightarrow \mathcal{Y}$. We consider the adaptation problem with $N$ source domains $\{\mathcal{D}_{s_{i}}\}_{i=1}^{N}$ and one target domain $\mathcal{D}_{t}$. In the $i$-th source domain $\mathcal{D}_{s_{i}}$, $X_{s_{i}}=\{x_{s_{i}}^k\}_{k=1}^{N_{s_{i}}}$ and $Y_{s_{i}}=\{y_{s_{i}}^k\}_{k=1}^{N_{s_{i}}}$ denote the observed data and corresponding labels sampled from the source distribution $P_{s_i}$, i.e., $(x_{s_{i}},y_{s_i})\sim P_{s_i}$. Let $X_{t}=\{x_{t}^k\}_{k=1}^{N_{t}}$ be the observed target data sampled from the target distribution $P_{t}$, and $Y_{t}=\{y_{t}^k\}_{k=1}^{N_{t}}$ be the \textit{unknown} target labels, i.e., $(x_{t},y_{t})\sim P_{t}$. In the setting of MDA, there are two basic assumptions: (1) samples from different domains share the same output space, i.e. $y_{s_{i}} \in \mathcal{Y}$, $y_{t} \in \mathcal{Y}$, and (2) domains have different distributions due to domain shift, i.e., $P_{s_{i}}\neq P_{s_{j}} $, $ P_{s_{i}}\neq P_{t}$, $\forall i,j\in \{1,2,\cdot\cdot\cdot,N\}$. The goal of MDA is to predict the labels of $X_{t}$ by exploiting the information of $\{(X_{s_{i}},Y_{s_{i}})\}_{i=1}^{N}$ and $X_{t}$.

\subsection{Pseudo Target Domain: Construction and Alignment}

\begin{figure}[t]
\centering{\includegraphics[scale=0.15]{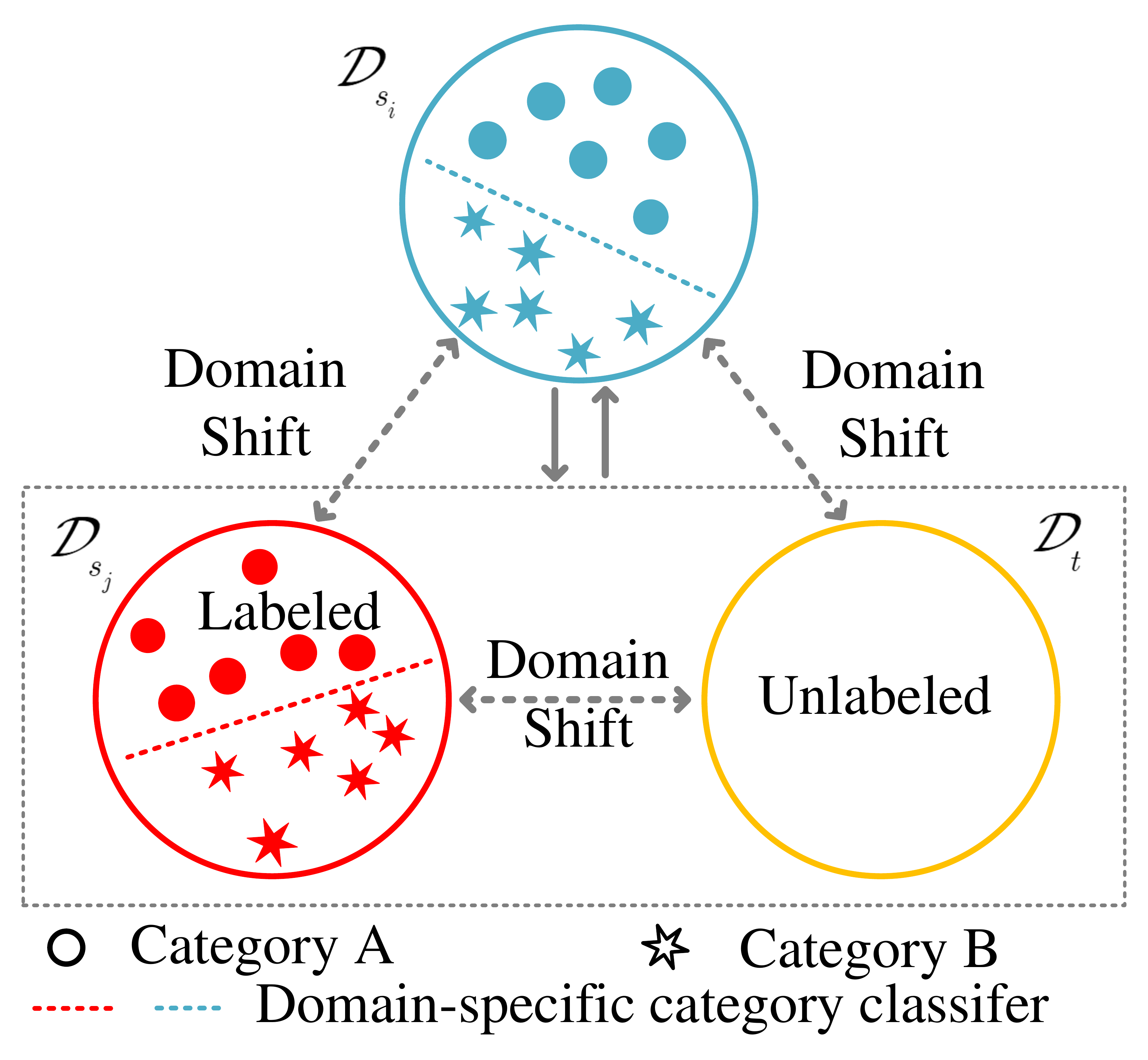}
\caption{Sketch map of PTMDA. Both $\mathcal{D}_{s_j}$ and $\mathcal{D}_t$ act as different yet related domains with respect to $\mathcal{D}_{s_i}$, while $\mathcal{D}_{s_j}$ has ground-truth label and $\mathcal{D}_t$ has none. It motivates us to construct a pseudo target domain which contains both label data from $\mathcal{D}_{s_j}$ and unlabeled data from $\mathcal{D}_t$. This approach renders $\mathcal{D}_t$ more discriminative or structured information, thus, it makes the alignment between $\mathcal{D}_t$ and $\mathcal{D}_{s_i}$ more effective. Better viewed in color.}\label{fig2}}
\end{figure}

Our method not only aims at reducing the domain shift between each group of source and target domains, it also seeks to extract discriminative knowledge from multiple sources to enrich the target domain. Without loss of generality, we consider $\mathcal{D}_{s_{i}}$, $\mathcal{D}_{s_{j}}$, and $\mathcal{D}_{t}$ as an example, where $i, j \in \{1,\cdot\cdot\cdot,N\}$ and $i \neq j$. The basic motivation of pseudo target domain construction has been shown in Fig.~\ref{fig2}. Generally, in domain adaptation, one can regularize the knowledge transfer model to the unlabeled target samples with available labeled source samples. It motivates us to mimic a new target domain, which is called \textit{pseudo target domain} in this work, to include both labeled data from $\mathcal{D}_{s_{j}}$ and unlabeled data from $\mathcal{D}_{t}$. In this way, the labeled data in the pseudo target domain can provide reliable supervision for network feature learning. The reasonability of pseudo target domain construction can be summarized as follows.
\begin{itemize}
  \item Domain shift exists not only between $\mathcal{D}_{t}$ and $\mathcal{D}_{s_{i}}$ but also across $\mathcal{D}_{s_{i}}$ and $\mathcal{D}_{s_{j}}$, thus, it is natural to treat $\mathcal{D}_{s_{j}}$ as a general target of $\mathcal{D}_{s_{i}}$, and $\mathcal{D}_{s_{j}}$ can be used to mimic a new and enlarged (with $\mathcal{D}_t$) target domain w.r.t. $\mathcal{D}_{s_i}$.
  \item In the alignment of target and source domains, samples from two domains are mapped to a shared space and drew closer. Consequently, it is rational to combining $\mathcal{D}_{s_{j}}$ and $\mathcal{D}_{t}$ after the alignment, which leads to the construction of the new pseudo target domain.
\end{itemize}

In summary, the PTMDA method is consisted of two stages. 1) For each pair of source domain $\mathcal{D}_{s_{j}}$ and target domain $\mathcal{D}_{t}$, we initialize a deep network using the adversarial training strategy with a metric constraint to reduce domain shift, and then combine the two mapped domains into a pseudo target domain $\hat{\mathcal{D}}_{s_{j},t}$. Samples in $\mathcal{D}_{t}$ are equipped with pseudo-labels in this stage. 2) To utilize discriminative and structured information across multiple source domains, each pseudo target domain generated in the first step is aligned to each of the remainder source domains separately.

In these process, each pair of source and target domains is chosen in turn to construct the pseudo target domain and then fed into the second stage. Thus, while domain difference between source and the target in the first stage has been minimized and has little effect on the second stage relatively, the second stage can take turns to align among source domains and utilize diverse distributed information from different sources.

\begin{figure}[tb]
\centering\includegraphics[scale=0.42]{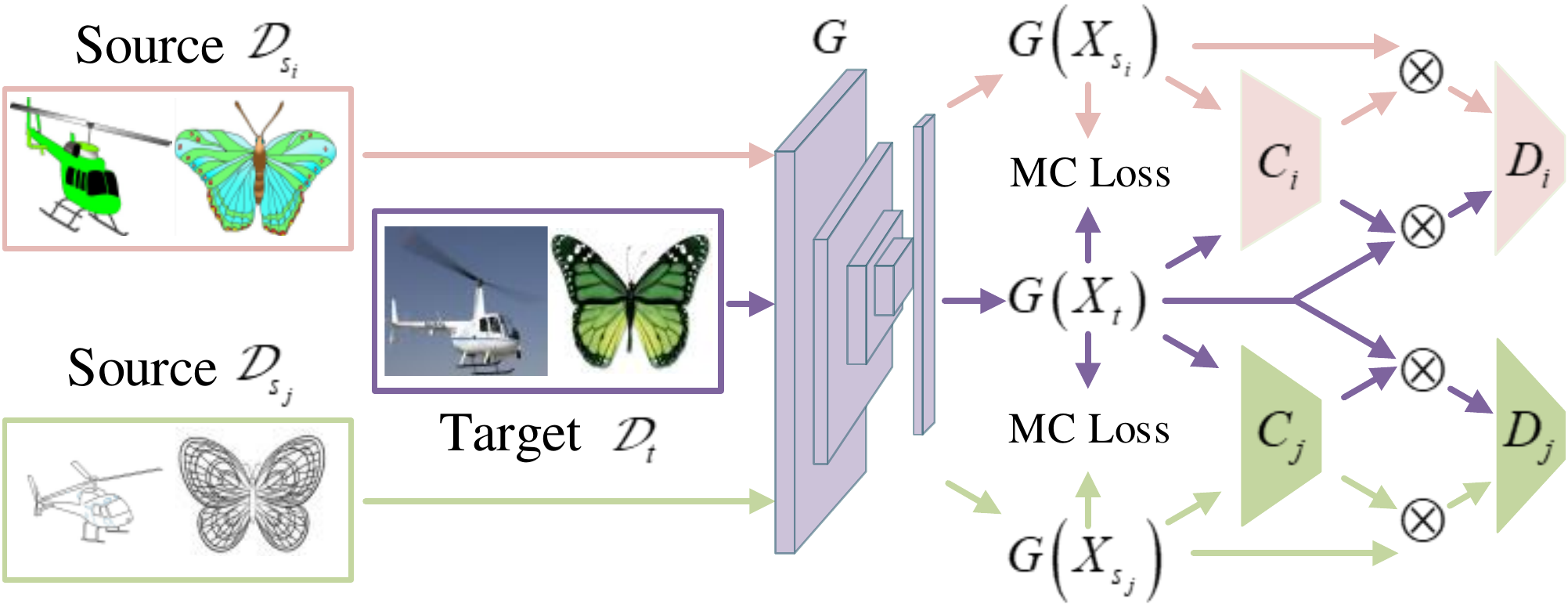}
\caption{Pseudo target domain construction. The feature extractor $G$ maps $\{\mathcal{D}_{s_{i}},\mathcal{D}_{t} \}$, $\{\mathcal{D}_{s_{j}},\mathcal{D}_{t}\}$ into two group-specific subspaces in turn. PTMDA reduces the domain shift within $\{\mathcal{D}_{s_{i}},\mathcal{D}_{t}\}$ and $ \{\mathcal{D}_{s_{j}},\mathcal{D}_{t}\}$, respectively.}\label{fig3}
\end{figure}

1) \textit{Stage 1: Pseudo Target Domain Construction.} In this stage, the adversarial training manner is used to align the distributions of each pair of source and target domains as much as possible. Then, the pair of domains are combined as a pseudo target domain, consisting of samples from both labeled source domain and pseudo-labeled target domain.

The model consists of a feature extraction network $G$ with parameters $\theta_{G}$, $N$ category classifiers $\{C_{i}\}_{i=1}^{N}$ with parameters $\{\theta_{C_{i}}\}_{i=1}^{N}$, and $N$ domain discriminators $\{D_{i}\}_{i=1}^{N}$ with parameters $\{\theta_{D_{i}}\}_{i=1}^{N}$. Each group of $\{G,C_{i},D_{i}\}$ addresses a batch of samples drawn from $\{(X_{s_{i}}, Y_{s_{i}}),X_{t}\}$.

As illustrated in Fig.~\ref{fig3}, each group of source domain and target domain (e.g., $\{\mathcal{D}_{s_{i}},\mathcal{D}_{t} \}$ or $ \{\mathcal{D}_{s_{j}},\mathcal{D}_{t} \}$) is used to construct a pseudo target domain. Specifically, for the group of $\{\mathcal{D}_{s_{i}},\mathcal{D}_{t} \}$, the feature extractor $G$ maps $\mathcal{D}_{s_{i}}$ and $\mathcal{D}_{t}$ into a shared latent space. It learns the feature mappings from different domains with a single generator network. The classifier $C_{i}$ predicts the category for the input samples from $\{\mathcal{D}_{s_{i}},\mathcal{D}_{t} \}$, and the domain discriminator $D_{i}$ supervises the learning process of $G$ towards the direction which learns domain invariant features with respect to $\{\mathcal{D}_{s_{i}},\mathcal{D}_{t} \}$.

In the training of feature extractor $G$ and discriminator $D_{i}$, the feature distributions of $\{(X_{s_{j}}, Y_{s_{j}})\}_{j=1}^{N}$ and $X_{t}$ are aligned using the following adversarial loss, i.e.,
\begin{eqnarray}\label{eq1}
 \nonumber L_{{adv}_{j}}(\theta_{G}, \theta_{D_{j}})\!\!&\!\!=\!\!&\!\!-\mathbb{E}_{x_{s_{j}} \sim P_{s_{j}}}\![\log (1\!-\!D_{j}(\varphi(G(x_{s_{j}}\!),\tilde{y}_{s_{j}}\!)))] \\
   \!\!&\!\! \!\!&\!\! - \mathbb{E}_{x_{t} \sim P_{t}}[\log (D_j(\varphi(G(x_{t}),\tilde{y}_{t})))].
\end{eqnarray}
Here $\tilde{y}_{s_j}$ and $\tilde{y}_{t}$ represent the category prediction probability for samples $x_{s_j}$ and $x_{t}$, respectively. $\varphi$ is a conditioning operator, which adds constraints to features $G(x_{s_{j}})$ and $G(x_{t})$ via $\tilde{y}_{s_{j}}$ and $\tilde{y}_{t}$, respectively. Let $d_{\mathrm{f}}$ and $d_{\mathrm{p}}$ be the dimensions of the input vectors $\mathrm{f}$ and $\mathrm{p}$, respectively. $d_{0}$ is a threshold. $\Pi_{\otimes}$ is the outer product of multiple vectors, and $\Pi_{\odot}$ is the explicit randomized multi-linear map. $\varphi$ is defined as
\begin{equation*}
\mathrm{\varphi(f,p)}=\left\{\begin{array}{ll}
\Pi_{\otimes}(\mathrm{f}, \mathrm{p}) & \text { if } d_{\mathrm{f}} \times d_{\mathrm{p}} \leq d_{0} \\
\Pi_{\odot}(\mathrm{f}, \mathrm{p}) & \text { otherwise }.
\end{array}\right.
\end{equation*}
To keep the computation efficiency of conditioning, the outer product on $\Pi_{\otimes}$ is approximated by the inner-product on $\Pi_{\odot}$ when the dimension of joint variable is larger than $d_{0}$. In~\cite{cdan1,mdd2020pami1}, $d_{0}$ is set to 4096, which is the largest feature dimension in typical DNNs, e.g., AlexNet. We use the same setting for simplicity and fairness. Details of $\varphi$ are depicted in \cite{cdan1}.

We also train the category classifier $C_{j}$ with the cross-entropy loss to preserve the essential discriminative capacity of the features, i.e.,
\begin{equation}\label{eq3}
L_{{cls}_{j}}(\theta_{G}, \theta_{C_{j}})\!=\!-\mathbb{E}_{(x_{s_{j}},y_{s_{j}}\!) \sim P_{s_{j}}}[y_{s_{j}}\log (C_{j}(F(x_{s_{j}})))].
\end{equation}

Note that the \textit{equilibrium challenge} often appears in the adversarial training manner~\cite{2017gan,mdd2020pami1,Fei2019DUAE,cdan1}. It means that adversarial training can potentially deteriorate discriminative structure of the input data. In other words, the feature distributions may not be well aligned even if the domain discriminator is fully confused. To alleviate this challenge, Pandey et al. \cite{GenUD2021} used a metric transformation to keep the source samples clustering near the corresponding categories in the feature space. Kim et al. \cite{kim2021selfreg} used a self-supervised contrastive loss to make the representations of the positive pair samples close. However, these interesting works are designed for the task of domain generalization, in which the target domain is unaccessible during training, thus, they rarely consider transferring discriminative information for the target domain.

In this work, we introduce a novel metric constraint (MC) loss into the adversarial learning process. Inspired by the Fisher Linear Discriminant Analysis\cite{RenDos2015}, which aims to maximize separation among distinct classes and minimize within-class variance, we add an additional constraint on the adversarial training process using the available category information. To be specific, assume that $G(x_{s_{j}}^{m})$ is the output of the last full connection layer of the feature extractor $G$ for sample $x_{s_{j}}^{m}$, $y_{s_{j}}^{m}$ is the corresponding category label. A normalization factor is first computed as
\begin{equation*}
\label{eqPT}
T_j=\frac{1}{B_{j}} \sum\limits_{\substack{m,n \in \{1,\cdots,B_j\}}} \| G(x_{s_{j}}^{m})-G(x_{s_{j}}^{n})\|_2^{2}.
\end{equation*}
Then, the MC loss is formulated as
\begin{equation}
\begin{split}
\label{eq4}
&L_{mc_{j}}(\theta_{G})= \\&  \mathbb{E}_{(x_{s_{j}},y_{s_{j}}) \sim P_{s_{j}}} \log \frac{\underset{{y_{s_{j}}^{m} \neq y_{s_{j}}^{n}}}\sum \exp(-\| G(x_{s_{j}}^{m})-G(x_{s_{j}}^{n})\|_2^2/T_j)}
{ \underset{{y_{s_{j}}^{m} = y_{s_{j}}^{n}}}\sum \exp(-\| G(x_{s_{j}}^{m})-G(x_{s_{j}}^{n})\|_2^2/T_j)},
\end{split}
\end{equation}
where $m, n\in \{1,\cdots,B_j\}$ and $m\neq n$. With this formulation, the MC loss preserves structured information from original data through the adversarial training manner, which ensures the discriminability of representation and thus is capable to improve the performance of adversarial learning.

The loss terms in Eqs. (\ref{eq1})-(\ref{eq4}) enforce each source distribution $P_{s_{j}}$ to align with the target distribution $P_{t}$. For each pair of source domain $\mathcal{D}_{s_{j}}$ and target domain $\mathcal{D}_{t}$, the feature extractor $G$ maps their features into a shared subspace, in which we get the pseudo target domain containing mapped samples of $\mathcal{D}_{s_{j}}$ and $\mathcal{D}_{t}$. To guarantee that the relationships among categories are preserved across source and target, we assign pseudo labels $\hat{Y}_{t}$ for the target samples using $G$ and $\{C_{j}\}_{j=1}^{N}$. The pseudo labels are obtained by the average of predictions from $N$ classifiers $\{C_{j}\}_{j=1}^{N}$. We denote the pseudo target domain as $\hat{\mathcal{D}}_{s_{j},t} $ with data $\{(X_{s_{j},t}, Y_{s_{j},t})\}$, which contains data $\{(X_{s_{j}}, Y_{s_{j}})\}$ and $\{(X_{t},\hat{Y}_{t})\}$, and the corresponding distribution is $\hat{P}_{s_{j},t }$. It is worth noting that not all the pseudo labels are correct. To ensure the credibility of those unlabeled samples in $\hat{\mathcal{D}}_{s_{j},t}$, we select the target samples with a confidence threshold $\kappa$. The selection rule is formulated as $\{x\in X_{t}|\max(\frac{1}{N}\sum_{i=1}^{N}C_{i}(x))>\kappa\}$.

We align each pair of source domain and target domain in turn, and get a number of pseudo target domains $\{\hat{\mathcal{D}}_{s_{1},t} ,\cdots,\hat{\mathcal{D}}_{s_{N},t}\}$, which corresponds to distributions $\{\hat{P}_{s_{1},t} ,\cdots,\hat{P}_{s_{N},t}\}$.

\begin{figure}[ht]
\centering\includegraphics[scale=0.35]{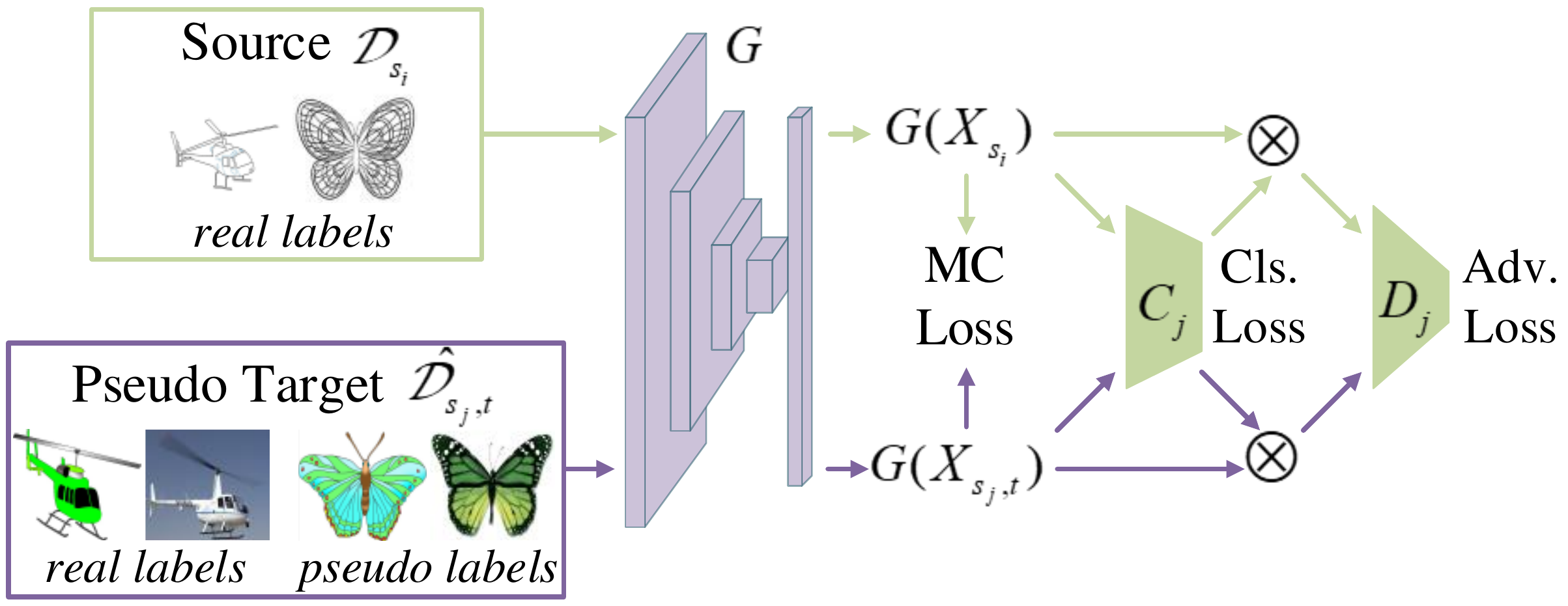}
\caption{Demonstration for aligning the remainder sources with the pseudo target. We take $\mathcal{D}_{s_i}$ and $\hat{\mathcal{D}}_{s_j,t}$ as an example. This strategy not only mitigates domain shift between $\{\mathcal{D}_t, \mathcal{D}_{s_i}\}$ and $\{\mathcal{D}_{s_j}, \mathcal{D}_{s_i}\}$, respectively, but also further improve alignment between $\mathcal{D}_t$ and$\mathcal{D}_{s_i}$ with the incorporation of $\mathcal{D}_{s_j}$.}\label{fig4}
\end{figure}

2) \textit{Stage 2: Aligning the Remainder Source Domains with the Pseudo Target.}  In this stage, we treat the MDA task as a set of single-source domain adaptation tasks. A similar adversarial training strategy is adopted to align the remainder source domains with the pseudo target domain in each subspace. The training process for stage 2 is illustrated in Fig. \ref{fig4}. We also take the $i$-th remainder source domain $\mathcal{D}_{s_{i}}$ and the $j$-th pseudo target domain $\hat{\mathcal{D}}_{s_{j},t}$ as an example, in which $\mathcal{D}_{s_{i}}$ needs to be aligned with $\hat{\mathcal{D}}_{s_{j},t}$. The architecture and parameters of network in stage 1 are shared in stage 2. It is worth noting that, this strategy not only mitigates domain shift between $\{ \mathcal{D}_{t}, \mathcal{D}_{s_i} \}$ and $\{ \mathcal{D}_{s_j}, \mathcal{D}_{s_i} \}$ , respectively, but also further improve alignment between $\mathcal{D}_{t}$ and $\mathcal{D}_{s_i}$ with the incorporation of $\mathcal{D}_{s_j}$.

For each pair of the source domain $\mathcal{D}_{s_{i}}$ and the pseudo target domain $\hat{\mathcal{D}}_{s_{j},t}$, the classification loss is written as
\begin{eqnarray}\label{eq5}
\nonumber L'_{{cls}_{j}}(\theta_{G}, \theta_{C_{j}})\!\!&\!\!=\!\!&\!\!-\mathbb{E}_{(x_{s_{i}}, y_{s_{i}} ) \sim P_{s_{i}}}[y_{s_{i}}\!\log (C_j(F(x_{s_i}\!))] \\
\!\!&\!\!\!\!&\!\!-\mathbb{E}_{(x_{s_{j}} ,y_{s_{j}}) \sim P_{s_{j}}}[y_{s_{j}}\!\log (C_j(F(x_{s_j}\!))].
\end{eqnarray}
We use the following adversarial loss to align the feature distributions of $\mathcal{D}_{s_{i}}$ and $\hat{\mathcal{D}}_{s_{j},t}$, i.e.,
\begin{eqnarray}\label{eq6}
\nonumber L'_{{adv}_{j}}(\theta_{G}, \theta_{D_{j}})\!\!\!&\!\!\!=\!\!\!&\!\!\!-\mathbb{E}_{x_{s_{i}} \sim P_{s_{i}}}[\log (1\!-\!D_{j}(\varphi(G(x_{s_{i}}),\tilde{y}_{s_{i}}\!))] \\
\!\!\!&\!\!\!\!\!\!&\!\!\!-\mathbb{E}_{x_{s_j,t}\sim\hat{P}_{s_j,t}}\![\log D_j(\varphi(G(x_{s_j,t}\!),\tilde{y}_{s_j,t}\!)],
\end{eqnarray}
where $x_{s_{j},t}$ represents sample from the $j$-th pseudo target domain $\hat{\mathcal{D}}_{s_{j},t}$, and $\tilde{y}_{s_{j},t}$ denotes the corresponding category prediction probability. The MC loss is
\begin{scriptsize}
\begin{equation}
\begin{split}
\label{eq7}
&L'_{mc_{j}}(\theta_{G})=\\& \mathbb{E}_{(x_{s_i},y_{s_i}) \sim P_{s_i}}\log\frac{\underset{{y_{s_i}^{m} \neq y_{s_i}^{n}}}\sum\!\!\exp(-\| G(x_{s_i}^{m})\!-\!G(x_{s_i}^{n})\|_2^2/T_i)}
{ \underset{{y_{s_i}^m = y_{s_i}^n}}\sum\!\!\exp(-\| G(x_{s_i}^m)\!-\!G(x_{s_i}^n)\|_2^2/T_i)} \\
& + \mathbb{E}_{(x_{s_j,t},y_{s_j,t}) \sim \hat{P}_{s_j,t}}\log\frac{\underset{{y_{s_j,t}^{m} \neq y_{s_j,t}^n}}\sum\!\!\exp(-\| G(x_{s_j,t}^m)\!-\!G(x_{s_j,t}^n)\|_2^2/T_{s_j,t})}
{\underset{{y_{s_j,t}^m = y_{s_j,t}^n}}\sum\!\exp(-\|G(x_{s_j,t}^m)\!-\!G(x_{s_j,t}^n)\|_2^2/T_{s_j,t})}.
\end{split}
\end{equation}
\end{scriptsize}

Each pair of source domain $\mathcal{D}_{s_{i}}$ and the pseudo target domain $\hat{\mathcal{D}}_{s_{j},t}$ is used in turn to supervise model training. This alternate learning manner will enhance the robustness towards domain shift and benefit the prediction for the target domain.

\subsection{Matching Normalization Layer}

The essence of MDA is the domain shift among diverse domains, so alleviating the impact of domain shift between each pair of source and target domains is the fundamental manipulation for the MDA task. Although DNNs excel at generalizing knowledge from the source domain to the target domain, the transferability of the latent layers of DNNs is still worthy of attention. In many UDA methods, feature extractors are constructed by using Batch Normalization (BN)~\cite{2015Batch,2018How}, which normalizes features from source and target domains by sharing the same pairs of mean and variance. This is sub-optimal due to the substantial differences existing between the distributions of diverse domains. The effectiveness of normalization would be probably deteriorated due to omitting this discordancy. Li et al.~\cite{AdaBN2018pr} used source samples to learn the BN parameters and used target samples to compute mean and variance for the test stage. Though the problem of discordancy is avoided, target samples make little contribution to the learnable affine parameters. Chang et al.~\cite{DSBN2019} proposed a two-stage method based on domain-specific batch normalization (DSBN) in DNNs. However, DSBN aims at separating domain-specific information from each domain, and it updates the affine parameters by using both domains.

In this section, we propose a matching normalization (MN) layer, which uses the same affine parameters for each pair of source and target domains with the gradient information from the target domain, to improve the performance of MDA. The adaptation performance would be improved by detaching domain-specific characteristic from the shared one. Based on this idea, we estimate the mean and variance of activations for source and target samples separately, and then normalize the feature map of each domain with their statistics respectively within the same batch and the same channel. The whitened features are expected to diminish domain-specific information. We denote a mini-batch of data as $\{x_{t}^{j}\}_{j=1}^{B_{t}} \bigcup \{x_{s_{i}}^{j}\}_{j=1}^{B_{i}}$, and corresponding network activation outputs as $\{h^{j}_{t}\}_{j=1}^{B_{t}} \bigcup \{h^{j}_{s_{i}}\}_{j=1}^{B_{i}}$, where $B_{t}$ and $B_{i}$ are the batch-size for the target domain and the $i$-th source domain, respectively. We estimate the sample mean and the sample variance for each channel, then normalize activation outputs (denoted as $\hat{h}^{j}_{t}$ and $\hat{h}^{j}_{s_{i}}$ for target domain and the $i$-th source domain, respectively) to have zero mean and unit variance. As a result, the feature maps of diverse domains are normalized to have similar distributions with the same mean and variance.

To preserve the network capacity, we only use the data flow in target domain to update the affine parameters $\gamma_{t}$ and $\beta_{t}$, i.e.,
\begin{eqnarray*}
\hat{z}^{j}_{t} &\triangleq& \mathrm{MN}_{\gamma_{t},\beta_{t}}(h^{j}_{t}) = \gamma_{t} \hat{h}^{j}_{t}+\beta_{t}, \\
\hat{z}^{j}_{s_{i}} &\triangleq& \mathrm{MN}_{\gamma_{t},\beta_{t}}(h^{j}_{s_{i}}) = \gamma_{t} \hat{h}^{j}_{s_{i}}+\beta_{t}.
\end{eqnarray*}

In light of the fact that discriminative information is usually dominated by the labeled source samples, the above training process can relieve over-fitting by sharing the affine parameters $\gamma_{t}$ and $\beta_{t}$. Besides, this strategy can align the distributions of the whitened source features with that of target features, and thus improve the transferability of the whole network.

\begin{figure}[htbp]
\subfigure[BN module]{\label{bn1}
\scalebox{0.53}{ 
\includegraphics{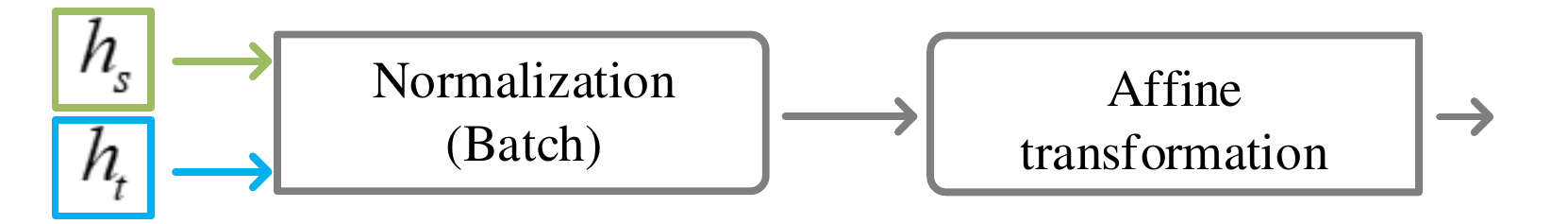}}}
\subfigure[MN module]{\label{mn1}
\scalebox{0.53}{ 
\includegraphics{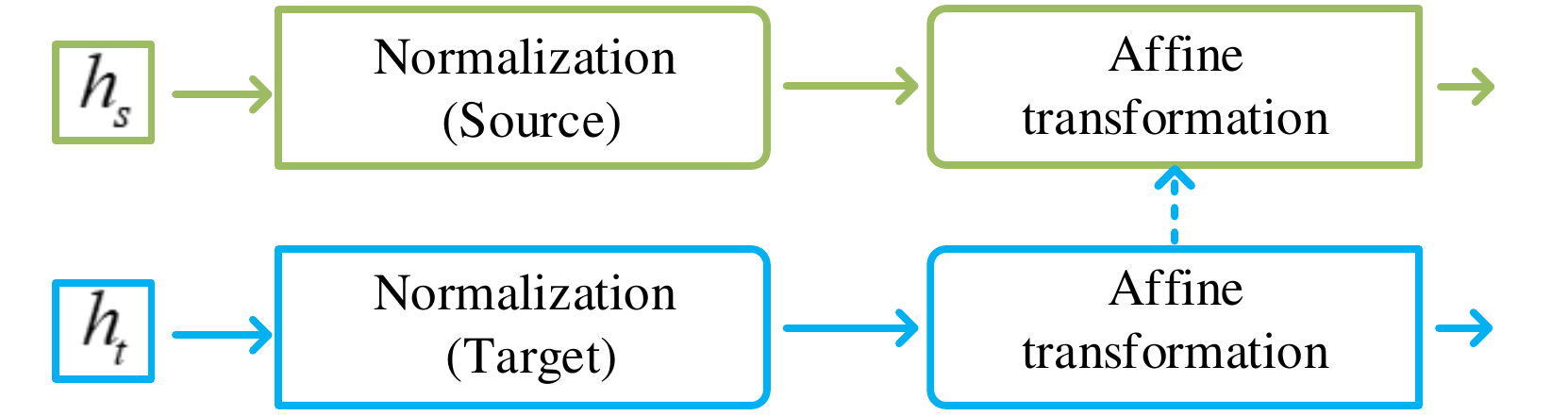}}}
\caption{Architectures of BN (a) and MN (b). $h_{s}$ and $h_{t}$ represent the activation of latent layers used in source and target domain, respectively.}\label{fig5}
\end{figure}

Main differences between MN and BN are shown in Fig. \ref{fig5}, and they can be summarized as follows.
\begin{enumerate}
\item{Each MN layer consists of two branches. One is for the source activations, and the other is for the target ones. }
\item{Each activation output in MN is normalized by the domain-specific statistics, which are estimated from those activation outputs in each domain.}
\item{The affine parameters are updated with the gradient information from the target domain, in comparison to BN which uses all data in the same batch.}
\end{enumerate}

\textit{Complexity Analysis.}  We consider a 4D tensor with dimension of $B\times C\times H\times W$, where $B$, $C$, $H$ and $W$ indicate batch-size, number of input channels, height and width of the input feature maps in one channel, respectively. For a batch of $B$ samples, MN estimates the statistics for each of the $H\times W$ pixels in the feature maps within $C$ channels respectively, and the computational complexity of MN is $\mathcal{O}(B C H W)$, which is comparable with BN. In other words, there is no increase in the parameter scale for MN.

The proposed MN layer is a generic component and can be plugged into many domain adaptation methods that use DNNs as the feature extractor. In this work, we replace each BN layer in feature extractor $G$ with the MN layer.

\subsection{Model Training}

The PTMDA algorithm is optimized as follows. For each group of $\mathcal{D}_{s_{i}}$ and $\mathcal{D}_{t}$,
\begin{equation}
\label{eq8}
\underset{\theta_{G}, \theta_{C_{i}}}\max\underset{\theta_{D_{i}}}\min~\lambda(L_{adv_{i}}(\theta_{G}, \theta_{D_{i}})\!-\!L_{mc_{i}}(\theta_{G}))\! -\! L_{cls_{i}}(\theta_{G}, \theta_{C_{i}}),
\end{equation}
then, for each pair of $\mathcal{D}_{s_{r}}$ and $\mathcal{D}_{s_{i},t}$,
\begin{equation}
\label{eq9}
\underset{\theta_G, \theta_{C_i}}\max\underset{\theta_{D_i}}\min~\lambda'(L'_{adv_i}(\theta_G, \theta_{D_i})\!-\!L'_{mc_i}\!(\theta_G)) \!-\! L'_{cls_i}\!(\theta_G, \theta_{C_i}),
\end{equation}
where $\lambda$ and $\lambda'$ are non-negative trade-off weights. 

We use the mini-batch stochastic gradient descent (Mini-batch SGD)~\cite{miniSGD2014} to perform standard back propagation optimization and solve the objective functions in Eq.~\eqref{eq8}-\eqref{eq9}. The domain discriminators $\{D_{i}\}_{i=1}^{N}$ and category classifiers $\{C_{i}\}_{i=1}^{N}$ are initialized with xavier\cite{xavier2010}. The detailed training procedure is shown in Algorithm~\ref{alg:PTMDA}.

\begin{algorithm}[ht]
	\caption{Pseudo-code for \textbf{PTMDA} method.}\label{alg:PTMDA}
    \begin{algorithmic}[1]
      \REQUIRE{Source sets $\{(X_{s_{i}}, Y_{s_{i}})\}_{i=1}^{N}$, target set $X_{t}$, parameters $\lambda$ and $\lambda'$, confidence threshold $\kappa$.}
     \ENSURE{Feature extractor $G$, domain discriminators $\{D_{i}\}_{i=1}^{N}$, category classifiers $\{C_{i}\}_{i=1}^{N}$.}\\
		\STATE Initialize $G$ with the model pretrained on ImageNet, and replace its BN layers with the MN layers. Initialize $\{D_{i}\}_{i=1}^{N}$ and $\{C_{i}\}_{i=1}^{N}$ with xavier.
        \FOR{$i=1$ to $N$}
        \STATE{Sample a mini-batch from $(X_{s_i},\!Y_{s_i})$ and $X_t$;}
        \STATE{Update $D_i$, $G$ and $C_i$ via Eq. (\ref{eq8});}
        \ENDFOR
        \STATE{Predict labels $\hat{Y}_{t}$ and select confidential samples for $X_{t}$.}
        \FOR{$i=1$ to $N$}
        \FOR{$j=1$ to $N$ and $j \neq i$}
        \STATE{Sample a mini-batch from $(X_{s_i}, Y_{s_i})$, $(X_{s_j}, Y_{s_j})$, and $(X_t,\hat{Y}_t)$;}
        \STATE{Update $D_{i}$, $G$ and $C_i$ via Eq. (\ref{eq9});}
        \ENDFOR
        \ENDFOR
	\end{algorithmic}
\end{algorithm}

\section{Theoretical Analysis}\label{sectionT}

In this section, we present some theoretical analysis for the effectiveness of PTMDA.

Ben-David et al.\cite{Ben2010MacLen} have proposed the following generalization bound under the MDA settings. Let $\mathcal{H}$ be a hypothesis space of VC dimension $d$. Given a total of $m$ labeled samples from all source domains. For each $j\in \{1,\cdots,N\}$, let $S_j$ be a labeled sample set of size $\beta_j$, with $\sum_{j=1}^N\beta_j\!=\!1$. Samples in $S_j$ are drawn from $\mathcal{D}_{s_j}$ and labeled via the labeling function $f_j$. $\epsilon_j(h)$ and $\epsilon_T(h)$ represent the errors of the hypothesis $h \in \mathcal{H}$ in source domain $\mathcal{D}_{s_{j}}$ and target domain $\mathcal{D}_{t}$, respectively. $\hat{\epsilon}_j(h)$ and $\hat{\epsilon}_T(h)$ are empirical errors. For any weight vector $\alpha \in \mathbb{R}_{+}^{N}$ with $\sum_{j=1}^N\alpha_{j}\!=\!1$, let $\hat{\epsilon}_{\alpha}(h)$ be the weighted error of some fixed hypothesis $h \in \mathcal{H}$ and $\hat{\epsilon}_{\alpha}(h)\!=\!\sum^N_{j=1}\alpha_j \hat{\epsilon}_j(h)$. If $\hat{h} \in \mathcal{H}$ is the empirical minimizer of $\hat{\epsilon}_{\alpha}(h)$ and $h_{T}^{*}=\min_{h \in \mathcal{H}}\epsilon_{T}(h)$ is the target error minimizer, then for any $\delta \in (0,1)$, with probability at least $1-\delta$,
\begin{eqnarray}\label{eq:MDAbound1}
  \nonumber \epsilon_{T}(\hat{h}) \!\!&\!\!\leq\!\!&\!\! \epsilon_{T}\left(h_{T}^{*}\right)+\sum_{j=1}^{N} \alpha_{j}d_{\mathcal{H} \Delta \mathcal{H}}(\mathcal{D}_{s_{j}}, \mathcal{D}_{t}) \\
   \!\!& \!\!+ \!\!& \!\! 2\sum_{j=1}^{N} \alpha_{j} \lambda_{j}\!+\!2\sqrt{\frac{d \log(2m)-\log\delta}{2 m}\sum_{j=1}^{N} \frac{\alpha_{j}^{2}}{\beta_{j}}},
\end{eqnarray}
where $\lambda_{j}=\min_{h \in \mathcal{H}}\{\epsilon_{T}(h)+\epsilon_{j}(h)\}$.

Using the triangle inequality, we have
\begin{eqnarray}\label{eq:MDAbound2}
\nonumber\epsilon_{T}(\hat{h}) \!\!&\!\!\leq \!\!&\!\! \epsilon_{T}\left(h_{T}^{*}\right)\!+\!\sum_{j=1}^{N}\!\alpha_{j}(d_{\mathcal{H} \Delta \mathcal{H}}(\mathcal{D}_{s_{j}}, \mathcal{D}^*_{s_{j}})\!+\!d_{\mathcal{H} \Delta \mathcal{H}}(\mathcal{D}^*_{s_{j}}, \mathcal{D}_{t})) \\
\!\!& \!\!+ \!\!& \!\! 2\sum_{j=1}^{N} \alpha_{j} \lambda_{j}\!+\!2\sqrt{\frac{d \log(2m)-\log\delta}{2 m}\sum_{j=1}^{N} \frac{\alpha_{j}^{2}}{\beta_{j}}},
\end{eqnarray}
where $\mathcal{D}^*_{s_{j}}$ is another source domain.
Once we fix the hypothesis class $\mathcal{H}$, the last two terms in Eq.~\eqref{eq:MDAbound1} and Eq.~\eqref{eq:MDAbound2} will be constant. We set $\alpha_{j}=1/N$ without loss of generality. Ganin et al.~\cite{DANN2015} explained that the optimal domain discriminator, which was explored in the adversarial training strategy, gives an upper bound for $d_{\mathcal{H} \Delta \mathcal{H}}(\mathcal{D}_{s_{j}}, \mathcal{D}_{t})$. Therefore, optimization of the adversarial loss $L_{{adv}_{j}}(\theta_{G}, \theta_{D_{j}})$ in Eq. (\ref{eq1}) actually minimizes an upper bound for $\sum^N_{j=1} \alpha_{j} d_{\mathcal{H} \Delta \mathcal{H}}(\mathcal{D}^*_{s_{j}}, \mathcal{D}_{t})$ when constructing the pseudo target domain, and optimization of the adversarial loss $L'_{{adv}_{j}}(\theta_{G}, \theta_{D_{j}})$ in Eq. (\ref{eq6}) minimizes an upper bound for $\sum^N_{j=1} \alpha_{j} d_{\mathcal{H} \Delta \mathcal{H}}(\mathcal{D}_{s_{j}}, \mathcal{D}^*_{s_{j}})$ when aligning the remainder source domains with the pseudo target domain. The first term in the generalization bound is approximately minimized by the classification loss $L_{{cls}_{j}}(\theta_{G}, \theta_{C_{j}})$ in Eq. (\ref{eq3}) and $L'_{{cls}_{j}}(\theta_{G}, \theta_{C_{j}})$ in Eq. (\ref{eq5}).

Due to the analysis above, PTMDA can reduce the target error bound in Eq. (\ref{eq:MDAbound2}) via minimizing the domain discrepancy not only between each pair of the source domains but also between each pair of the source and target domains. Therefore, PTMDA leads to a better approximation of the target risk, and works well in MDA settings.

\section{Experimental Results}\label{section4}

In this section, we first introduce the datasets and some experimental details. Then, we compare the proposed MN with BN on the single-source UDA tasks, and evaluate PTMDA on image classification tasks under MDA settings. At last, feature visualization, ablation study and other analysis are also presented.

\subsection{Datasets and Experimental Details}

\begin{table*}[htbp]
\small
  \caption{Accuracy (\%) comparison between MN and other batch normalization methods on Office-31.}
  \centering
  \renewcommand{\tabcolsep}{0.6pc} 
  \renewcommand{\arraystretch}{1.1} 
    \begin{tabular}{cccccccc}
    \hline
    Method & A$\rightarrow$W  & D$\rightarrow$W  & W$\rightarrow$D  & A$\rightarrow$D  & D$\rightarrow$A  & W$\rightarrow$A  & Avg \\
    \hline\hline
    CDAN+BN & 94.1$\pm$0.1  & 98.6$\pm$0.1  & 100$\pm$0.0   & 92.9$\pm$0.2  & 71.0$\pm$0.3  & 69.3$\pm$0.3  & 87.7 \\
    \hline
    CDAN+DSBN\cite{DSBN2019} & 91.2$\pm$0.1  & 100$\pm$0.0  & 100$\pm$0.0   & 91.5$\pm$0.2  & 68.8$\pm$0.4  & 67.4$\pm$0.2  & 86.5$\pm$0.1 \\
    \hline
    CDAN+DA-layer\cite{BoostMda2018}  & 94.0$\pm$0.3  & 99.0$\pm$0.1  & 100$\pm$0.0   & 95.2$\pm$0.1  & 70.4$\pm$0.2  & 72.2$\pm$0.3  & 88.4$\pm$0.1 \\
    \hline
    CDAN+MultiDIAL\cite{MultiDial2020}  & 91.5$\pm$0.3  & 99.3$\pm$0.2  & 100$\pm$0.0   & 92.0$\pm$0.5  & 71.5$\pm$0.3  & 72.3$\pm$0.3  & 87.7$\pm$0.2 \\
    \hline
    \textbf{CDAN+MN} &\textbf{94.8$\pm$0.1} & 99.2$\pm$0.2 & 100$\pm$0.0 & \textbf{93.9$\pm$0.2} & \textbf{74.0$\pm$0.4} & \textbf{75.1$\pm$0.3} & \textbf{89.5$\pm$0.1} \\
    \hline\hline
    BSP+BN \cite{Bsp2019} & 93.3$\pm$0.2  & 98.2$\pm$0.2  & 100$\pm$0.0   & 93.0$\pm$0.2  & 73.6$\pm$0.3  & 72.6$\pm$0.3  & 88.5 \\
    \hline
    \textbf{BSP+MN} & \textbf{96.1$\pm$0.1}  & \textbf{100$\pm$0.0}  & \textbf{100$\pm$0.0}   & \textbf{95.6$\pm$0.1} & 73.5$\pm$0.3  & \textbf{73.0$\pm$0.4}  & \textbf{89.7$\pm$0.1} \\
    \hline\hline
    SymNets+BN\cite{Kui2019CVPR}  & 90.8$\pm$0.1  & 98.8$\pm$0.3  & 100$\pm$0.0   & 93.9$\pm$0.5  & 74.6$\pm$0.6 & 72.5$\pm$0.5  & 88.4 \\
    \hline
    \textbf{SymNets+MN}  & \textbf{93.8$\pm$0.4}  & \textbf{99.5$\pm$0.1}  & \textbf{100$\pm$0.0}   & 93.2$\pm$0.2  & \textbf{75.1$\pm$0.3}  & \textbf{75.3$\pm$0.5}  & \textbf{89.5$\pm$0.1} \\
    \hline
    \end{tabular}%
  \label{table1}%
\end{table*}

\begin{table*}[htbp]
  \small
  \caption{Accuracy (\%) comparison between MN and other batch normalization methods on ImageCLEF-DA.}
  \centering
  \renewcommand{\tabcolsep}{0.6pc} 
  \renewcommand{\arraystretch}{1.1} 
    \begin{tabular}{cccccccc}
    \hline
    Method & I$\rightarrow$P  & P$\rightarrow$I  & I$\rightarrow$C  & C$\rightarrow$I  & C$\rightarrow$P  & P$\rightarrow$C  & Avg \\
    \hline\hline
    CDAN+BN & 77.7$\pm$0.3  & 90.7$\pm$0.2  & 97.7$\pm$0.3  & 91.3$\pm$0.3  & 74.2$\pm$0.2  & 94.3$\pm$0.3  & 87.7 \\
    \hline
    CDAN+DSBN\cite{DSBN2019} & 79.0$\pm$0.2  & 92.3$\pm$0.3  & 96.5$\pm$0.2  & 86.3$\pm$0.3  & 75.3$\pm$0.2  & 94.1$\pm$0.1  & 87.3$\pm$0.1 \\
    \hline
    CDAN+DA-layer\cite{BoostMda2018} & 79.2$\pm$0.2  & 91.8$\pm$0.2  & 96.7$\pm$0.3  & 92.0$\pm$0.1  & 77.5$\pm$0.2  & 94.0$\pm$0.3  & 88.5$\pm$0.1 \\
    \hline
    CDAN+MultiDIAL\cite{MultiDial2020} & 78.3$\pm$0.2  & 92.3$\pm$0.2  & 97.8$\pm$0.3  & 92.8$\pm$0.2  & 78.5$\pm$0.4  & 94.7$\pm$0.4  & 89.1$\pm$0.1 \\
    \hline
    \textbf{CDAN+MN} & \textbf{80.0$\pm$0.3}  & \textbf{92.7$\pm$0.3}  & 97.3$\pm$0.1  & \textbf{93.0$\pm$0.2}  & \textbf{78.2$\pm$0.1}  & \textbf{95.2$\pm$0.1}  & \textbf{89.4$\pm$0.1} \\
    \hline\hline
    BSP+BN \cite{Bsp2019} & 79.6$\pm$0.2  & 91.8$\pm$0.2  & 95.8$\pm$0.2  & 92.8$\pm$0.4  & 77.0$\pm$0.2  & 94.5$\pm$0.3  & 88.9$\pm$0.1 \\
    \hline
    \textbf{BSP+MN} & \textbf{81.5$\pm$0.4}  & \textbf{92.1$\pm$0.2}  & \textbf{96.5$\pm$0.2}  & \textbf{93.8$\pm$0.4}  & \textbf{79.1$\pm$0.2}  & \textbf{95.4$\pm$0.3}  & \textbf{89.7$\pm$0.1} \\
    \hline\hline
    SymNets+BN\cite{Kui2019CVPR} & 80.2$\pm$0.3  & 93.6$\pm$0.2  & 97.0$\pm$0.3  & 93.4$\pm$0.3  & 78.7$\pm$0.3  & 96.4$\pm$0.1  & 89.9 \\
    \hline
    \textbf{SymNets+MN} & \textbf{81.8$\pm$0.3}  & 92.7$\pm$0.3  & 96.6$\pm$0.2 & \textbf{94.3$\pm$0.3}  & \textbf{79.4$\pm$0.4}  & \textbf{96.8$\pm$0.2}  & \textbf{90.3$\pm$0.1} \\
    \hline
    \end{tabular}%
  \label{table3}%
\end{table*}

We evaluate the PTMDA method on five benchmarks.

\textbf{Office31.} It is a collection of images from three different domains, i.e., Amazon, DSLR and Webcam~\cite{Saenko2010da1}. Each domain consists of 31 categories which are commonly encountered in family and office, such as bike, desk, and phone.

\textbf{Office-Caltech10.} It consists of a subset of the Office31 dataset with three domains and an additional Caltech domain, including 10 classes common to the four domains.

\textbf{ImageCLEF-DA.} It consists of subsets from Caltech-256 (C), ImageNet ILSVRC 2012 (I), and Pascal VOC 2012 (P). Each domain is comprised of 12 common categories with 50 images in each category.

\textbf{DomainNet.} It consists of six different domains, namely Clipart, Infograph, Painting, Quickdraw, Real, and Sketch\cite{Peng_2019_ICCV}. Each domain contains 345 categories of objects. Following the protocol of the VisDA2019 Challenge, we use the training and test splits of the given data for each domain.

\textbf{Digits-five.} It consists of five datasets, i.e., MNIST, MNIST-M~\cite{Yaroslav2016JMLR}, Synthetic Digits~\cite{Yaroslav2016JMLR}, SVHN, and USPS. Each dataset contains 10 categories of images. We use $mt$, $mm$, $sy$, $sv$ and $up$ to represent these domains for short, respectively.

Unless otherwise stated, the experiments are conducted on the MDA settings\footnote{In section~\ref{sect:experiments-MN}, which aims to show the superiority of MN over BN, we just perform single-source UDA experiments due to the time afford.}, and the proposed MN layers are used in PTMDA. We perform each task for five runs and report the average accuracy and standard deviation.

In this work, we follow the same experimental settings as M3SDA~\cite{Peng_2019_ICCV}. For each benchmark dataset, each domain is selected as the unlabeled target domain in turn, and the rest are used as labeled source domains. Taking the Digits-Five dataset as an example, we randomly sample 25,000 images from the training split and 9,000 images from the test split for MNIST, MNIST-M, Synthetic Digits, and SVHN. Since USPS only contains 9,298 images, we use the entire dataset as a separate domain. $mt$, $mm$, $sy$, $sv$ $\rightarrow$ $up$ means that the training splits of  $mt$, $mm$, $sy$ and $sv$ domains are used as the source domains, and the training split of $up$ is used as the target domain. Finally, we evaluate the PTMDA method on the test split of domain $up$. In the test phase of PTMDA, test data is fed into the feature extractor $G$ and then predicted with each classifier $C_{i}$, and final prediction is made by the average of predictions from $N$ classifiers.

We employ three fully-connected layers for each discriminator, and use a single fully-connected layer as the classifier. In the Digits-Five based experiments, we use three convolution layers and two fully-connected layers in the feature extractor. For the experiments on Office-31, Office-Caltech10, and ImageCLEF-DA, we use ResNet-50~\cite{resnet2016} pre-trained on ImageNet as the feature extractor. ResNet-101~\cite{resnet2016} is employed on the DomainNet dataset. We set the output dimension of the feature extractor as 512. All the network parameters are optimized by a Mini-batch SGD optimizer with the weight decay set as 5$e$-4, the momentum as 0.9, and the initial learning rate as 1$e$-3. The confidence threshold $\kappa$ can filter out those target samples with less confident (unreliable) pseudo labels, and it is set to 0.98 in this work. In order to suppress the noisy signals brought by domain discriminators and pseudo labels of target samples at the early training stages, we gradually change the hyper-parameters $\lambda$ and $\lambda'$ from 0 to 1 during the training procedure with the following schedule:
$$\lambda=\lambda'=\frac{2}{1+\exp (-10 \cdot \frac{epoch}{\#\{epoch\}}) }-1,$$ where $epoch$ represents the current epoch index and $\#\{epoch\}$ is the total size of epochs.

Following the evaluation protocol extensively employed in existing MDA works \cite{Peng_2019_ICCV,Xu_2018_CVPR}, we introduce the following two standards:
\begin{itemize}
\item \textbf{Source-combine}: We combine images from all source domains as a single source domain and conduct single-source unsupervised domain adaptation task.
\item \textbf{Multi-source}: We conduct comparisons with existing MDA methods. We also compare PTMDA with several related single-source UDA algorithms, where single-source UDA models are trained on each pair of source and target domains, and then predictions on test data are combined.
\end{itemize}

From the experiments in source-combine scenario, we verify the necessity of developing MDA models. Comparing the performance of PTMDA with that of other MDA methods, we validate that PTMDA can aggregate information from multiple source domains effectively. The setting of source-only is used as a baseline, where all images from source domains are used to train a classifier without considering the target domain.

\subsection{Experiments for MN}\label{sect:experiments-MN}

We design our MN layer in a generic way and make it a plug-and-play alternate for the BN layer without additional modification to the network architecture. To demonstrate the efficiency and effectiveness of the MN layer, we first perform single-source UDA experiments on three datasets with MN, and compare the results with those of BN. CDAN \cite{cdan1} is a popular method recently proposed for UDA, and its network architecture contains BN layers. So we choose CDAN as the baseline. For fair comparisons, we follow the experimental settings by Long et al.~\cite{cdan1}, while replacing the BN layers with MN layers in the CDAN framework. We use ResNet-50\cite{resnet2016} as the backbone network. The batch size is set to 36. We implement all the experiments in the PyTorch library and use an NVIDIA GeForce TITAN Xp GPU.

We conduct experiments on the Office-31 and ImageCLEF-DA datasets, and show the results in Table \ref{table1} and Table \ref{table3}, respectively. MN achieves the best results on both datasets with clear margins. Especially for the Office-31, MN surpasses BN with an improvement of 1.8\%. This confirms that MN can improve the transferability of DNNs and significantly boost the performance of existing domain adaptation methods. Although DSBN\cite{DSBN2019}, DA-layer\cite{BoostMda2018} and MultiDIAL\cite{MultiDial2020} are recent UDA-aware batch normalization methods, MN outperforms all of them. These results confirm that MN is a potential alternative to the original BN and the domain-specific BN.

To further verify that the proposed MN layer can be plugged into other domain adaptation methods, we also combine MN with two recent domain adaptation methods, i.e., BSP~\cite{Bsp2019} and SymNets~\cite{Kui2019CVPR}. The experimental results are also summarized in Table~\ref{table1} and \ref{table3}. Compared with original results using BN, both the performance of BSP and SymNets are improved when using the proposed MN layer. This confirms that MN is a generic component and can boost the performance of other domain adaptation methods.

\subsection{MDA Results on Office-Caltech10}

\begin{table*}[!htb]
\small
\caption{Accuracy(\%) comparison among recent MDA methods on Office-Caltech10 with ResNet-50.}
\centering
\renewcommand{\tabcolsep}{0.8pc} 
\renewcommand{\arraystretch}{1.1} 
\begin{tabular}{ccccccc}
\hline
    & Method & A,C,D$\rightarrow$W & A,C,W$\rightarrow$D & A,D,W$\rightarrow$C & C,D,W$\rightarrow$A & Avg \\
\hline
\multirow{2}{*}{Source-combine}
&Source-only & 99.0 & 98.3 & 87.8 & 86.1 & 92.8 \\
\cline{2-7}
&DAN\cite{Long2019PAMI1}         & 99.3 & 98.2 & 89.7 & 94.8 & 95.5 \\
\cline{2-7}
&DANN\cite{Yaroslav2016JMLR} & 96.5 & 99.1 & 89.2 & 94.7 & 94.8 \\
\cline{2-7}
&DSBN\cite{DSBN2019}    & 98.8$\pm$0.2 & 99.9$\pm$0.1 & 94.6$\pm$0.1 & 92.5$\pm$0.1 & 96.4$\pm$0.1 \\
\cline{2-7}
&DSAN\cite{DSAN2020}    & 99.6$\pm$0.3 & 99.2$\pm$0.1 & 91.3$\pm$0.2 & 92.7$\pm$0.4 & 95.7$\pm$0.2 \\
\hline
\multirow{8}{*}{Multi-source}
&Source-only & 99.1 & 98.2 & 89.7 & 94.8 & 95.5 \\
\cline{2-7}
&DAN\cite{Long2019PAMI1}         & 99.5 & 99.1 & 89.2 & 91.6 & 94.8 \\
\cline{2-7}
&DANN\cite{Yaroslav2016JMLR} & 99.4$\pm$0.2 & 96.5$\pm$0.5 & 91.2$\pm$0.3 & 93.2$\pm$0.1 & 95.1$\pm$0.2 \\
\cline{2-7}
&D-CORAL\cite{coral2016AAAI} & 99.3$\pm$0.1 & 98.9$\pm$0.1 & 91.0$\pm$0.2 & 93.2$\pm$0.1 & 95.6$\pm$0.1\\
\cline{2-7}
&JAN\cite{LongM2017jda} & 99.4 & 99.4 & 91.2 & 91.8 & 95.5 \\
\cline{2-7}
&MEDA\cite{Jindong2018meda} & 99.3 & 99.2 & 91.4 & 92.9 & 95.7 \\
\cline{2-7}
&MCD\cite{Saito2018CVPRmcd}  & 99.5 & 99.1 & 91.5 & 92.1 & 95.6 \\
\cline{2-7}
&DCTN\cite{Xu_2018_CVPR} & 99.4 & 99.0 & 90.2 & 92.7 & 95.3 \\
\cline{2-7}
&M3SDA\cite{Peng_2019_ICCV} & 99.5 & 99.2 & 92.2 & 94.5 & 96.4 \\
\cline{2-7}
&MFSAN\cite{zhu2019MSDA} & 99.7 & 99.4 & 93.8 & 95.4 & 97.1 \\
\cline{2-7}
&\textbf{PTMDA} & \textbf{100.0$\pm$0.0} & \textbf{100.0$\pm$0.0} & \textbf{96.5$\pm$0.2} & \textbf{96.7$\pm$0.4} & \textbf{98.3$\pm$0.1} \\
\hline
\end{tabular}
\label{table6}%
\end{table*}

Comparisons on classification performance between PTMDA and the state-of-the-art approaches on Office-Caltech10 datasets are show in Table \ref{table6}. We can see that most methods have very high accuracy when $D$ or $W$ is used as the target domain, while PTMDA yields superior performance in all cases due to its ability to leverage structured information among source domains to promote generalization. As compared with the Source-combine scenario, PTMDA exceeds the accuracy of the Source-only model and DAN by $5.5 \%$ and $2.8 \%$, respectively. The results verify that it is effectual to develop algorithm for MDA task rather than simple combination of diverse source domains. In the multi-source scenario, we compare PTMDA with several state-of-the-art approaches. For single-source UDA methods, model is trained on different source domains and the predictions are combined for the testing data. For DAN, DANN and JAN, which use the adversarial training strategy to match the joint feature distribution cross different domains, PTMDA outperforms them by $3.5\%$, $3.2\%$ and $2.8\%$, respectively. In regard to those MDA algorithms, the average performance of PTMDA exceeds the baseline DCTN, M3SDA, and MFSAN, by $3\%$, $1.9\%$, and $1.2\%$, respectively. These experimental results indicate that PTMDA can extract more transferable features from various source domains than state-of-the-art approaches.

\begin{table}[!htb]
\small
\caption{Accuracy(\%) comparison among recent MDA methods on ImageCLEF-DA with ResNet-50.}
\centering
\renewcommand{\tabcolsep}{0.15pc} 
\renewcommand{\arraystretch}{1.1} 
\begin{tabular}{cccccc}
\hline
    & Method  & I,C $\rightarrow $ P & I,P $\rightarrow $ C & P,C $\rightarrow $ I & Avg \\
\hline
\multirow{5}{1.2cm}{Source-combine}
&Source-only & 77.2 & 92.3 & 88.1 & 85.8 \\
\cline{2-6}
&DAN\cite{Long2019PAMI1} & 77.6 & 93.3 & 92.2 & 87.7 \\
\cline{2-6}
&ADDA\cite{Tzeng2017CVPRadda}         & 76.5 & 94.0 & 93.2 & 87.0 \\
\cline{2-6}
&DANN\cite{Yaroslav2016JMLR}    & 77.9 & 93.7 & 91.8 & 87.8 \\
\cline{2-6}
&D-CORAL\cite{coral2016AAAI}    & 77.1 & 93.6 & 91.7 & 87.5 \\
\cline{2-6}
&DSBN\cite{DSBN2019}    & 77.7$\pm$0.2 & 94.1$\pm$0.3 & 91.9$\pm$0.1 & 87.9$\pm$0.1 \\
\cline{2-6}
&DSAN\cite{DSAN2020}    & 77.6$\pm$0.2 & 95.1$\pm$0.1 & 91.4$\pm$0.6 & 88.1$\pm$0.3 \\
\hline
\multirow{3}{1.2cm}{Multi  -source}
&DANN\cite{Yaroslav2016JMLR}         & 74.5$\pm$0.4 & 93.7$\pm$0.5 & 87.8$\pm$0.3 & 85.4$\pm$0.2 \\
\cline{2-6}
&D-CORAL\cite{coral2016AAAI}         & 77.7$\pm$0.1 & 93.5$\pm$0.1 & 91.5$\pm$0.2 & 87.6$\pm$0.1 \\
\cline{2-6}
&DCTN\cite{Xu_2018_CVPR}         & 75.0 & 95.7 & 90.3 & 87.0 \\
\cline{2-6}
&MFSAN\cite{zhu2019MSDA} & 79.1 & 95.4 & 93.6 & 89.4 \\
\cline{2-6}
&\textbf{PTMDA} & \textbf{79.1$\pm$0.2} & \textbf{97.3$\pm$0.3} & \textbf{94.1$\pm$0.3} & \textbf{90.2$\pm$0.1} \\
\hline
\end{tabular}
\label{table8}%
\end{table}

\subsection{MDA Results on ImageCLEF-DA}

We also evaluate PTMDA on the more challenging ImageCLEF-DA datasets and summarize the classification accuracies in Table \ref{table8}. PTMDA outperforms several state-of-the-art methods on most tasks. Specifically, PTMDA achieves comparable accuracy on $I,C \rightarrow P$, and exceeds all the compared methods in other cases. The average accuracy of PTMDA on all the three transfer tasks is $90.2\%$, which increases by $0.8\%$ against the base competitor MFSAN. The performance of DCTN is inferior to that of the Source-combine setting, and it seems that the ensemble of adversarial-based classifiers tends to unstable when the target domain displays large domain shift among source domains. In addition, our proposed PTMDA outperforms DANN and D-CORAL which use the ensemble of multiple classifiers in the multi-source setting by a large margin on all adaptation tasks, which verifies the effectiveness of our method.

\subsection{MDA Results on Office31}

\begin{table}[!t]
\small
\caption{Accuracy(\%) comparison among recent MDA methods on Office31 with ResNet-50.}
\centering
\renewcommand{\tabcolsep}{0.06pc} 
\renewcommand{\arraystretch}{1.1} 
\begin{tabular}{cccccc}
\hline
    & Method &A,W$\rightarrow$D & A,D$\rightarrow$W & D,W$\rightarrow$A & Avg \\
\hline
\multirow{4}{1.2cm}{Source-combine}
&Source-only & 99.2 & 93.4 & 56.1 & 82.8 \\
\cline{2-6}
&DANN\cite{Yaroslav2016JMLR}     & 99.7 & 98.1 & 67.6 & 88.5 \\
\cline{2-6}
&DAN\cite{Long2019PAMI1}         & 99.6 & 97.8 & 67.6 & 88.3 \\
\cline{2-6}
&D-CORAL\cite{coral2016AAAI}         & 99.3 & 98.0 & 67.1 & 88.1 \\
\cline{2-6}
&DSBN\cite{DSBN2019}    & 99.0$\pm$0.2 & 98.8$\pm$0.2 & 70.1$\pm$0.3 & 89.3$\pm$0.1 \\
\cline{2-6}
&DSAN\cite{DSAN2020}    & 99.1$\pm$0.1 & 98.6$\pm$0.1 & 72.4$\pm$0.2 & 90.0$\pm$0.1 \\
\hline
\multirow{4}{1.2cm}{Multi -source}
&DANN\cite{Yaroslav2016JMLR}         & 99.1$\pm$0.1 & 98.3$\pm$0.2 & 73.3$\pm$0.3 & 90.2$\pm$0.2 \\
\cline{2-6}
&D-CORAL\cite{coral2016AAAI}         & 99.2$\pm$0.1 & 98.9$\pm$0.3 & 69.2$\pm$0.1 & 89.1$\pm$0.1 \\
\cline{2-6}
&DCTN\cite{Xu_2018_CVPR}         & 99.3 & 98.2 & 64.2 & 87.2 \\
\cline{2-6}
&MADAN\cite{2021MADAN} & 99.4 & 98.4 & 63.9 & 87.2 \\
\cline{2-6}
&Adv-Ensemble\cite{German1} & 99.3 & 97.3 & 68.1 & 88.3 \\
\cline{2-6}
&MFSAN\cite{zhu2019MSDA} & 99.5 & 98.5 & 72.7 & 90.2 \\
\cline{2-6}
&DSBN\cite{DSBN2019} & 100.0 & \textbf{99.9} & \textbf{75.6} & \textbf{91.8} \\
\cline{2-6}
&\textbf{PTMDA} & \textbf{100.0$\pm$0.0} & 99.6$\pm$0.2 & 75.4$\pm$0.4 & 91.7$\pm$0.1 \\
\hline
\end{tabular}
\label{table7}%
\end{table}
We report the experimental results on Office-31 based on ResNet-50 in Table \ref{table7}. It shows that the performance of DSBN\cite{DSBN2019} is slightly better than our PTMDA. The good performance of DSBN could probably due to the reason that it uses a semantic matching loss to align the centroids of the same classes across domains and achieves semantic transfer among diverse domains. PTMDA outperforms other compared methods on all tasks with an average classification accuracy of $91.7\%$. Transfer task on target domain $A$ is more challenging due to the large variations on resolution of the images between $A$ and other domains. With respect to this harder task, both DCTN and MADAN\cite{2021MADAN} are inferior to those single-source UDA algorithms in the Source-combine scenario, while PTMDA still exceeds most methods and attains an absolute accuracy improvement of $2.7\%$ against the latest MFSAN\cite{zhu2019MSDA}. Both DCTN and MADAN are based on adversarial learning, this result further testifies that it is necessary to add the metric constraint into the adversarial learning process.

\begin{table}[htbp]
\small
\caption{Accuracy(\%) comparison among recent MDA methods on Office31 with AlexNet.}
\centering
\renewcommand{\tabcolsep}{0.1pc} 
\renewcommand{\arraystretch}{1.1} 
\begin{tabular}{cccccc}
\hline
    & Method  &A,W$\rightarrow$D & A,D$\rightarrow$W & D,W$\rightarrow$A & Avg \\
\hline
\multirow{4}{1.2cm}{Source -combine}
&Source-only & 98.1 & 93.2 & 50.2 & 80.5 \\
\cline{2-6}
&DAN\cite{Long2019PAMI1}& 98.8 & 95.2 & 53.4 & 82.5 \\
\cline{2-6}
&DANN\cite{Yaroslav2016JMLR}& 98.8 & 96.2 & 54.6 & 83.2 \\
\cline{2-6}
&D-CORAL\cite{coral2016AAAI} & 98.8 & 94.4 & 53.5 & 82.3 \\
\cline{2-6}
&DSBN\cite{DSBN2019}& 99.0$\pm$0.0 & 95.1$\pm$0.0 & 51.3$\pm$0.2 & 81.8$\pm$0.1  \\
\cline{2-6}
&DSAN\cite{DSAN2020} & 99.1$\pm$0.1 & 93.6$\pm$0.1 & 50.4$\pm$0.2 & 81.0$\pm$0.1 \\
\hline
\multirow{4}{1.2cm}{Multi  -source}
&Source-only &98.2 & 92.7 & 51.6 & 80.8 \\
\cline{2-6}
&DANN\cite{Yaroslav2016JMLR}& 98.1$\pm$0.1 & 93.4$\pm$0.3 & 52.5$\pm$0.2 & 81.4$\pm$0.2 \\
\cline{2-6}
&D-CORAL\cite{coral2016AAAI} & 98.6$\pm$0.3 & 94.7$\pm$0.2 & 53.3$\pm$0.2 & 82.1$\pm$0.2 \\
\cline{2-6}
&DA-layer\cite{BoostMda2018} & 94.8 & 95.8 & \textbf{62.9} & 84.5 \\
\cline{2-6}
&LtC-MSDA\cite{Lcombine2020} & \textbf{99.6} & 97.2 & 56.9 & 84.6 \\
\cline{2-6}
&MultiDIAL\cite{MultiDial2020} & 97.2 & 95.3 & 62.7 & \textbf{85.1} \\
\cline{2-6}
&\textbf{PTMDA} &  99.4$\pm$0.1 & \textbf{97.3$\pm$0.2} & 53.5$\pm$0.2 & 83.4$\pm$0.1 \\
\hline
\end{tabular}
\label{table7alex}%
\end{table}

There are a number of existing MDA approaches using AlexNet as the backbone, as it has been used for a long time in this field. We also use the AlexNet as the backbone and compare our PTMDA with these methods on the Office31 dataset. Table \ref{table7alex} shows that DA-layers\cite{BoostMda2018}, LtC-MSDA\cite{Lcombine2020} and MultiDIAL\cite{MultiDial2020} outperform our PTMDA. DA-layers can discover latent domains and exploit this latent structure to learn a robust target classifier. LtC-MSDA constructs a knowledge graph on the prototypes of various domains to transfer information. MultiDIAL is designed not only to align the feature distributions among various domains but also to automatically decide the degree of alignment at different levels of the deep network. PTMDA outperforms DA-layers and MultiDIAL on the first two tasks. We also note that DA-layers and MultiDIAL outperform PTMDA on the third task which uses \textit{Amazon} as the target domain. It is worth noting that the transfer task on target domain \textit{Amazon} is more challenging than others, due to the large variations on resolution of the images between domain \textit{Amazon} and other domains. In this task, the amount of samples in the source domains is relatively small (i.e., 498 images in the \textit{DSLR} domain and 795 images in the \textit{Webcam} domain), while the target domain has 2,817 images. PTMDA performs alignment between \textit{DSLR} and \textit{Amazon}, as well as alignment between \textit{Webcam} and \textit{Amazon} during the training procedure. This setting may lead to over-fitting among the source domains which have few training samples.

In addition, PTMDA outperforms all the Source-combine-based methods in the average sense. It validates again the effectiveness of PTMDA in transferring knowledge from multiple source domains to target domain.

\subsection{MDA Results on DomainNet}

The results on the DomainNet benchmark are shown in Table \ref{table9domainnet}. We can see that PTMDA achieves comparable accuracy compared with these methods. Specifically, PTMDA outperforms all the compared methods when Clipart, Painting or Sketch is selected as the target domain, which verifies the effectiveness of PTMDA. PTMDA also obtains $0.7\%$, $3.0\%$ and $4.0\%$ absolute improvements compared with the recent MDA methods CMSS\cite{cmss2020}, SHOT\cite{shot2020} and MDDA\cite{DistillMSDA2020AAAI}. This is because CMSS, SHOT and MDDA focus only on aligning multiple source domains with the target, while PTMDA aims at reducing the domain shift which exists not only between source and target domains but also among diverse source domains. Surprisingly, most of the MDA methods achieve lower accuracy than the results of Source-combine scenario in the Quickdraw task. This can be explained by the fact that there are large domain gaps between Quickdraw and the other domains. While DCTN and MDDA also performs multi-way adversarial learning to address the shift between each source and target domain, our PTMDA achieves $9.0\%$ and $4.0\%$ performance improvements over them, which validates the effectiveness of the proposed MC loss term in adversarial learning. Moreover, PTMDA outperforms Source-combine methods which use the mixed data from multiple source domains. It indicates that exploiting structured information among diverse source domains could benefit the adaptation performance. In addition, our proposed PTMDA outperforms DANN and D-CORAL which use the ensemble of classifiers by a large margin in the multi-source setting on all adaptation tasks, which verifies the effectiveness of our method.

\begin{table*}[htbp]
\small
\caption{Classification accuracy(\%) on DomainNet based on ResNet-101. ``Clipart'' means that Clipart is target domain and the others are source domains.}
\centering
\renewcommand{\tabcolsep}{0.4pc} 
\renewcommand{\arraystretch}{1.1} 
\label{table4}
\resizebox{\textwidth}{!}{
\begin{tabular}{ccccccccc}
\hline
    & Method  &Clipart & Infograph & Painting & Quickdraw & Real & Sketch & Avg \\
\hline
\multirow{6}{*}{Source-combine}
&Source-only & 47.6$\pm$0.5 & 13.0$\pm$0.4 & 38.1$\pm$0.5 & 13.3$\pm$0.4 & 51.9$\pm$0.9 & 33.7$\pm$0.5 & 32.9$\pm$0.5 \\
\cline{2-9}
&DAN\cite{Long2019PAMI1}         & 45.4$\pm$0.5 & 12.8$\pm$0.9 & 36.2$\pm$0.6 & 15.3$\pm$0.4 & 48.6$\pm$0.7 & 34.0$\pm$0.5 & 32.1$\pm$0.6 \\
\cline{2-9}
&JAN\cite{LongM2017jda}         & 40.9$\pm$0.4 & 11.1$\pm$0.6 & 35.4$\pm$0.5 & 12.1$\pm$0.7 & 45.8$\pm$0.6 & 32.3$\pm$0.6 & 29.6$\pm$0.6 \\
\cline{2-9}
&DANN\cite{Yaroslav2016JMLR}        & 45.5$\pm$0.6 & 13.1$\pm$0.7 & 37.0$\pm$0.7 & 13.2$\pm$0.8 & 48.9$\pm$0.7 & 31.8$\pm$0.6 & 32.6$\pm$0.7 \\
\cline{2-9}
&ADDA\cite{Tzeng2017CVPRadda}        & 47.5$\pm$0.8 & 11.4$\pm$0.7 & 36.7$\pm$0.5 & 14.7$\pm$0.5 & 49.1$\pm$0.8 & 33.5$\pm$0.5 & 32.2$\pm$0.6 \\
\cline{2-9}
&MCD\cite{Saito2018CVPRmcd}         & 54.3$\pm$0.6 & 22.1$\pm$0.7 & 45.7$\pm$0.6 &  7.6$\pm$0.5 & 58.4$\pm$0.7 & 43.5$\pm$0.6 & 38.5$\pm$0.6 \\
\cline{2-9}
&DSBN\cite{DSBN2019}      & 45.5$\pm$0.5 & 19.3$\pm$0.1 & 45.5$\pm$0.1 &  6.7$\pm$0.2 & 54.6$\pm$0.4 & 36.6$\pm$0.2 & 34.7$\pm$0.1\\
\cline{2-9}
&DSAN\cite{DSAN2020}      & 53.4$\pm$0.3 & 20.1$\pm$0.5 & 40.5$\pm$0.2 &  14.6$\pm$0.2 & 57.4$\pm$0.1 & 45.2$\pm$0.1 & 38.5$\pm$0.2 \\
\hline
\multirow{3}{*}{Multi-source}
&DANN\cite{Yaroslav2016JMLR}        & 47.4$\pm$0.3 & 21.5$\pm$0.2 & 49.7$\pm$0.3 & 9.3$\pm$0.4 & 59.3$\pm$0.2 & 36.6$\pm$0.1 & 37.3$\pm$0.1 \\
\cline{2-9}
&D-CORAL\cite{coral2016AAAI}  & 49.5$\pm$0.2 & 24.7$\pm$0.2 & 53.5$\pm$0.2 & 11.5$\pm$0.3 & 59.6$\pm$0.3 & 46.6$\pm$0.2 & 40.9$\pm$0.1\\
\cline{2-9}
&DCTN\cite{Xu_2018_CVPR}        & 48.6$\pm$0.7 & 23.5$\pm$0.6 & 48.8$\pm$0.6 & 7.2$\pm$0.5 & 53.5$\pm$0.6 & 47.3$\pm$0.5 & 38.2$\pm$0.6 \\
\cline{2-9}
&M3SDA\cite{Peng_2019_ICCV}       & 58.6$\pm$0.5 & 26.0$\pm$0.9 & 52.3$\pm$0.6 &  6.3$\pm$0.6 & 62.7$\pm$0.5 & 49.5$\pm$0.8 & 42.6$\pm$0.6 \\
\cline{2-9}
&MDDA\cite{DistillMSDA2020AAAI}       & 59.4$\pm$0.6 & 23.8$\pm$0.8 & 53.2$\pm$0.6 &  12.5$\pm$0.6 & 61.8$\pm$0.5 & 48.6$\pm$0.8 & 43.2 \\
\cline{2-9}
&SHOT\cite{shot2020}       & 61.7 & 22.2 & 52.6 &  12.2 & 67.7 & 48.6 & 44.2 \\
\cline{2-9}
&CMSS\cite{cmss2020}       & 64.2$\pm$0.2 & 28.0$\pm$0.2 & 53.6$\pm$0.4 &  16.0$\pm$0.1 & 63.4$\pm$0.2 & 53.8$\pm$0.4 & 46.5$\pm$0.2 \\
\cline{2-9}
&LtC-MSDA\cite{Lcombine2020}       & 63.1$\pm$0.5 & \textbf{28.7$\pm$0.7} & 56.1$\pm$0.5 &  \textbf{16.3$\pm$0.5} & \textbf{66.1$\pm$0.6} & 53.8$\pm$0.6 & \textbf{47.4} \\
\cline{2-9}
&\textbf{PTMDA} & \textbf{66.0$\pm$0.3} & 28.5$\pm$0.2 & \textbf{58.4$\pm$0.4} & 13.0$\pm$0.5 & 63.0$\pm$0.24 & \textbf{54.1$\pm$0.3} & 47.2$\pm$0.1\\
\hline
\end{tabular}}
\label{table9domainnet}%
\end{table*}

\begin{figure}[tb]
\centering{
\subfigure[Source-only]{\label{Fig.sub.1-1}
\begin{minipage}[b]{0.15\textwidth} 
\centering \scalebox{0.165}{ 
\includegraphics{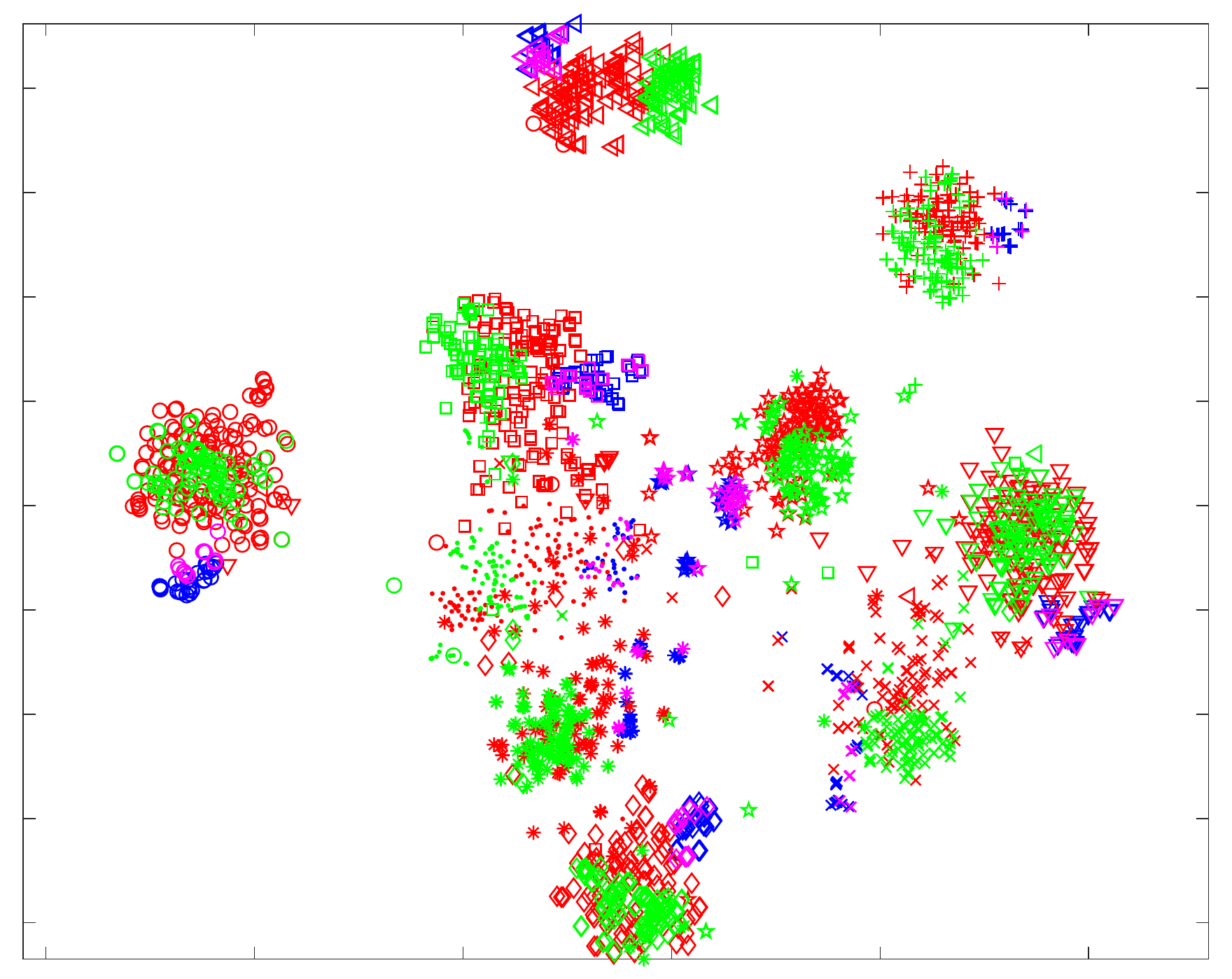}}
\end{minipage}}
\subfigure[M3SDA]{\label{Fig.sub.1-2}
\begin{minipage}[b]{0.15\textwidth}
\centering \scalebox{0.165}{ 
\includegraphics{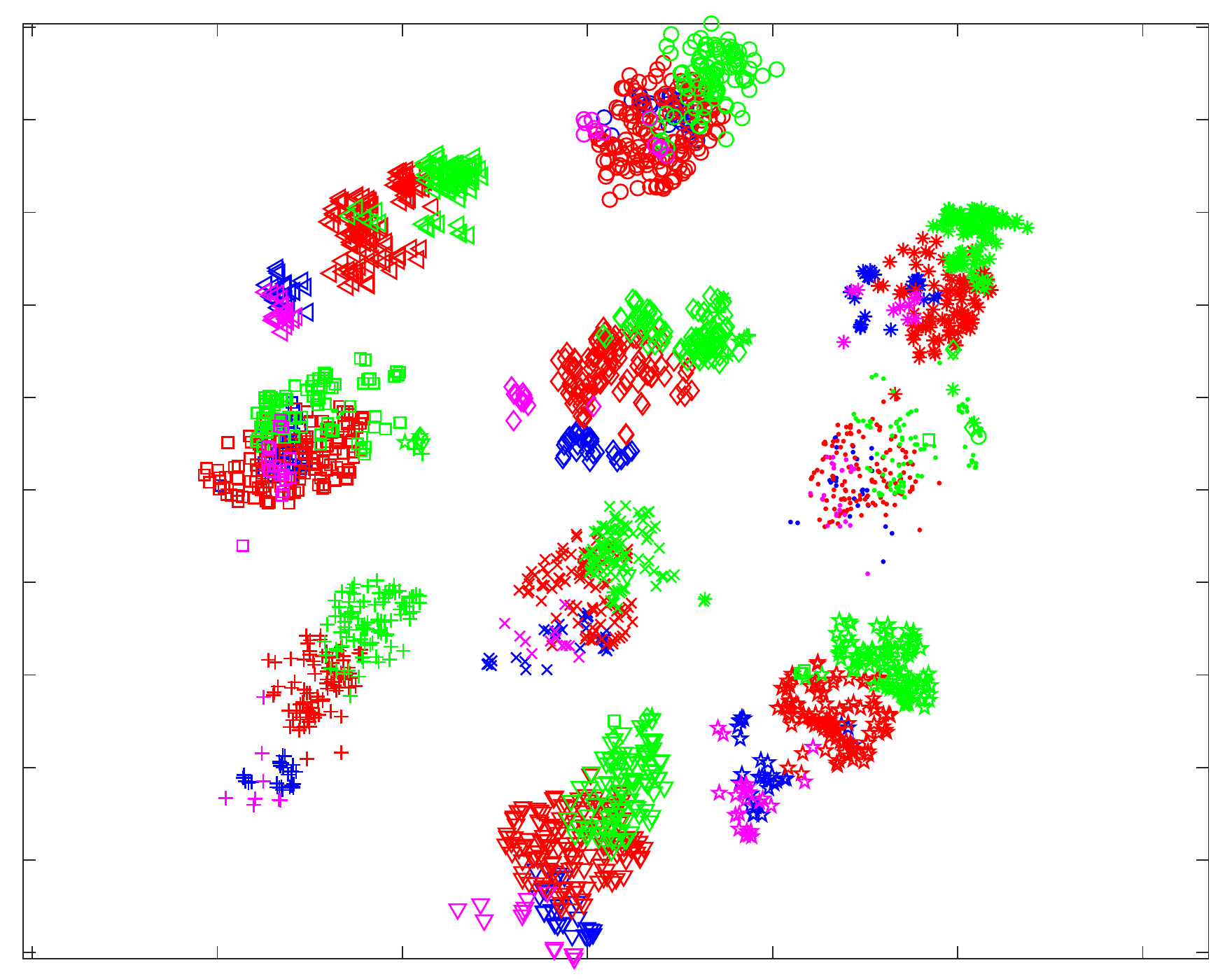}}
\end{minipage}}
\subfigure[PTMDA]{\label{Fig.sub.1-3}
\begin{minipage}[b]{0.15\textwidth}
\centering \scalebox{0.165}{ 
\includegraphics{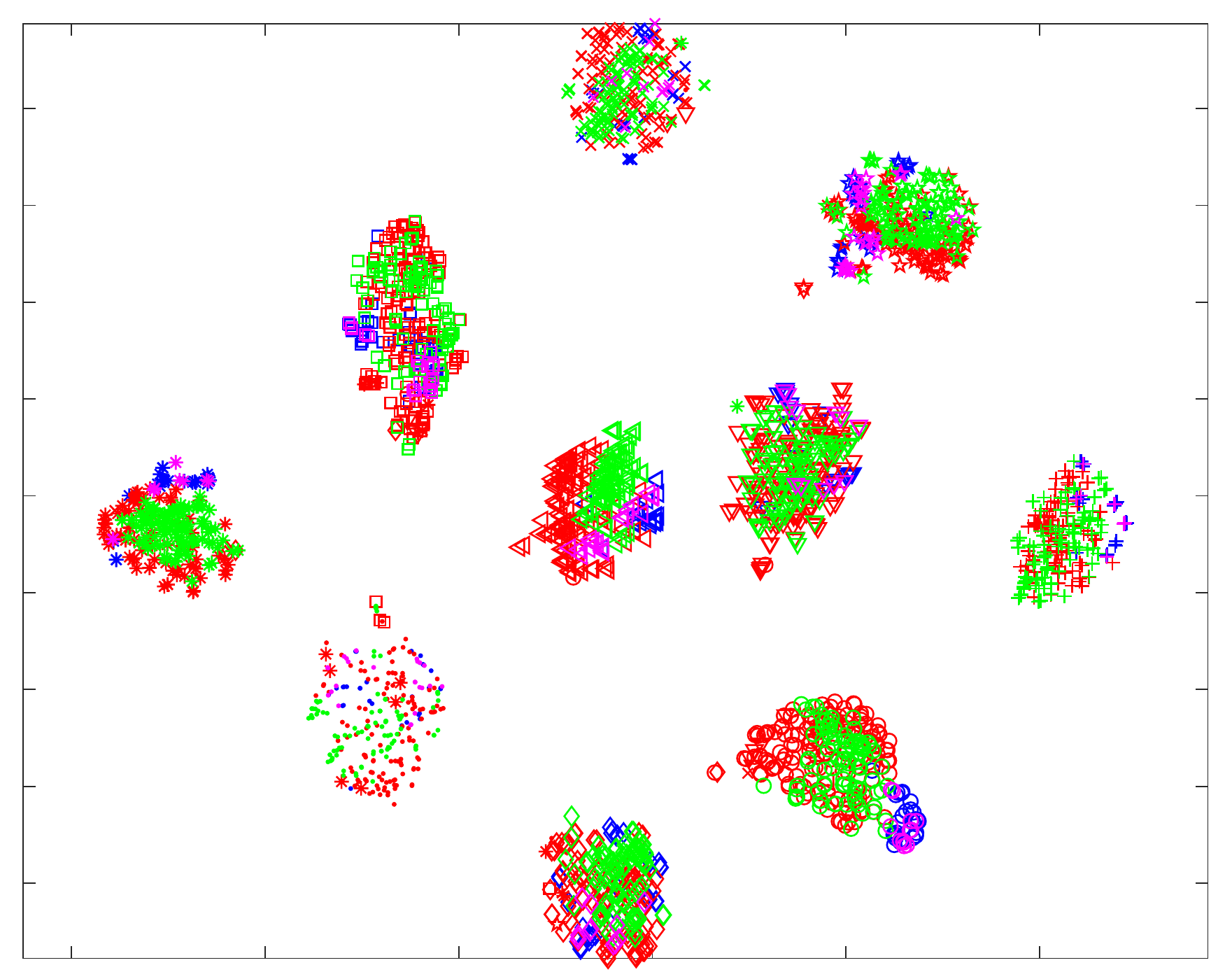}}
\end{minipage}}
\caption{Visualization on task $A,D,W \rightarrow C$. The target domain $C$ is colored in red, and the source domains are shown by other colors. Features in the same category are visualized with the same marker.}\label{fig7}}
\end{figure}

\subsection{MDA Results on Digits-five}

\begin{table*}[htbp]
\small
\caption{Accuracy(\%) comparison among recent MDA methods on Digits-five under full protocol \cite{Peng_2019_ICCV}.}
\centering
\renewcommand{\tabcolsep}{0.3pc} 
\renewcommand{\arraystretch}{1.1} 
\resizebox{\textwidth}{!}{
\begin{tabular}{cccccccc}
\hline
    & Method  &mt,up,sv,sy$\rightarrow$mm& mm,up,sv,sy$\rightarrow$mt & mm,mt,sv,sy$\rightarrow$up &  mm,mt,up,sy$\rightarrow$sv &mm,mt,up,sv$\rightarrow$sy & Avg \\
\hline
\multirow{3}{*}{Source-combine}
&Source-only & 63.7$\pm$0.8 & 92.3$\pm$0.9 & 90.7$\pm$0.5 & 71.5$\pm$0.8 & 83.4$\pm$0.8 & 80.3$\pm$0.8 \\
\cline{2-8}
&DAN\cite{Long2019PAMI1}         & 67.9$\pm$0.8 & 97.5$\pm$0.6 & 93.5$\pm$0.9 & 67.8$\pm$0.8 & 86.9$\pm$0.9 & 82.7$\pm$0.8 \\
\cline{2-8}
&DANN\cite{Yaroslav2016JMLR}        & 70.8$\pm$0.9 & 97.9$\pm$0.8 & 93.5$\pm$0.8 & 68.5$\pm$0.9 & 87.4$\pm$0.7 & 83.6$\pm$0.8 \\
\cline{2-8}
&DSBN\cite{DSBN2019}      & 68.6$\pm$0.1 & 96.3$\pm$0.2 & 93.5$\pm$0.2 & 75.4$\pm$0.1 & 86.5$\pm$0.1 & 84.0$\pm$0.1\\
\cline{2-8}
&DSAN\cite{DSAN2020}      & 78.1$\pm$0.3 & 96.4$\pm$0.3 &  92.3$\pm$0.2 & 76.4$\pm$0.2 & 87.8$\pm$0.2 & 86.2$\pm$0.2 \\
\hline
\multirow{11}{*}{Multi-source}
&Source-only & 63.4$\pm$0.7 & 90.5$\pm$0.8 & 88.7$\pm$0.9 & 63.5$\pm$0.9 & 82.4$\pm$0.7 & 77.7$\pm$0.8 \\
\cline{2-8}
&DAN\cite{Long2019PAMI1}         & 63.8$\pm$0.7 & 96.3$\pm$0.5 & 94.2$\pm$0.9 & 62.5$\pm$0.7 & 85.4$\pm$0.8 & 80.4$\pm$0.7 \\
\cline{2-8}
&DANN\cite{Yaroslav2016JMLR} & 71.3$\pm$0.6 & 97.6$\pm$0.8 & 92.3$\pm$0.9 & 63.5$\pm$0.8 & 85.3$\pm$0.8 & 82.0$\pm$0.8 \\
\cline{2-8}
&D-CORAL\cite{coral2016AAAI}  & 62.5$\pm$0.7 & 97.2$\pm$0.8 & 93.5$\pm$0.8 & 64.4$\pm$0.7 & 82.8$\pm$0.7 & 80.1$\pm$0.8 \\
\cline{2-8}
&JAN\cite{LongM2017jda} & 65.9$\pm$0.9 & 97.2$\pm$0.7 & 95.4$\pm$0.8 & 75.3$\pm$0.7 & 86.6$\pm$0.6 & 84.1$\pm$0.7 \\
\cline{2-8}
&ADDA\cite{Tzeng2017CVPRadda} & 71.6$\pm$0.5 & 97.9$\pm$0.8 & 92.8$\pm$0.7 & 75.5$\pm$0.5 & 86.5$\pm$0.6 & 84.8$\pm$0.6 \\
\cline{2-8}
&MEDA\cite{Jindong2018meda} & 71.3$\pm$0.8 & 96.5$\pm$0.8 & 97.0$\pm$0.8 & 78.5$\pm$0.8 & 84.6$\pm$0.8 & 85.6$\pm$0.8 \\
\cline{2-8}
&MCD\cite{Saito2018CVPRmcd}  & 72.5$\pm$0.7 & 96.2$\pm$0.8 & 95.3$\pm$0.7 & 78.9$\pm$0.8 & 87.5$\pm$0.7 & 86.1$\pm$0.7 \\
\cline{2-8}
&DCTN\cite{Xu_2018_CVPR} & 70.5$\pm$1.2 & 96.2$\pm$0.8 & 92.8$\pm$0.3 & 77.6$\pm$0.4 & 86.8$\pm$0.8 & 84.8$\pm$0.7 \\
\cline{2-8}
&M3SDA\cite{Peng_2019_ICCV} & 72.8$\pm$1.1 & 98.4$\pm$0.7 & 96.1$\pm$0.8 & 81.3$\pm$0.9 & 89.6$\pm$0.6 & 87.7$\pm$0.8 \\
\cline{2-8}
&ADAGE\cite{hallu2019}  & 85.3$\pm$0.2 & 98.3$\pm$0.3 & 97.1$\pm$0.3 & \textbf{85.3$\pm$0.2} & \textbf{96.2$\pm$0.1} & \textbf{92.4} \\
\cline{2-8}
&MDDA\cite{DistillMSDA2020AAAI} & 78.6 & 98.8 & 93.9 & 79.3 & 89.7 & 88.1 \\
\cline{2-8}
&SHOT\cite{shot2020} & 80.2$\pm$0.4 & 98.2$\pm$0.4 & 97.1$\pm$0.3 & 84.5$\pm$0.3 & 91.1$\pm$0.2 & 90.2 \\
\cline{2-8}
&CMSS\cite{cmss2020} & 75.3$\pm$0.6 & 99.0$\pm$0.1 & 97.7$\pm$0.1 & 88.4$\pm$0.5 & 93.7$\pm$0.2 & 90.8$\pm$0.3 \\
\cline{2-8}
&LtC-MSDA\cite{Lcombine2020}  & \textbf{85.6$\pm$0.8} & 99.0$\pm$0.4 & \textbf{98.3$\pm$0.4} & 83.2$\pm$0.6 & 93.0$\pm$0.5 & 91.8 \\
\cline{2-8}
&\textbf{PTMDA} & 85.2$\pm$0.4 &  \textbf{99.3$\pm$0.1} & 97.6$\pm$0.3 & 82.5$\pm$0.5 & 93.4$\pm$0.3 & 91.6$\pm$0.2 \\
\hline
\end{tabular}}
\label{table5}%
\end{table*}

\begin{figure}[tb]
\centering{
\subfigure[The actual target domain $C$]{\label{TSNE3-1}
\begin{minipage}[b]{0.23\textwidth} 
\centering \scalebox{0.25}{ 
\includegraphics{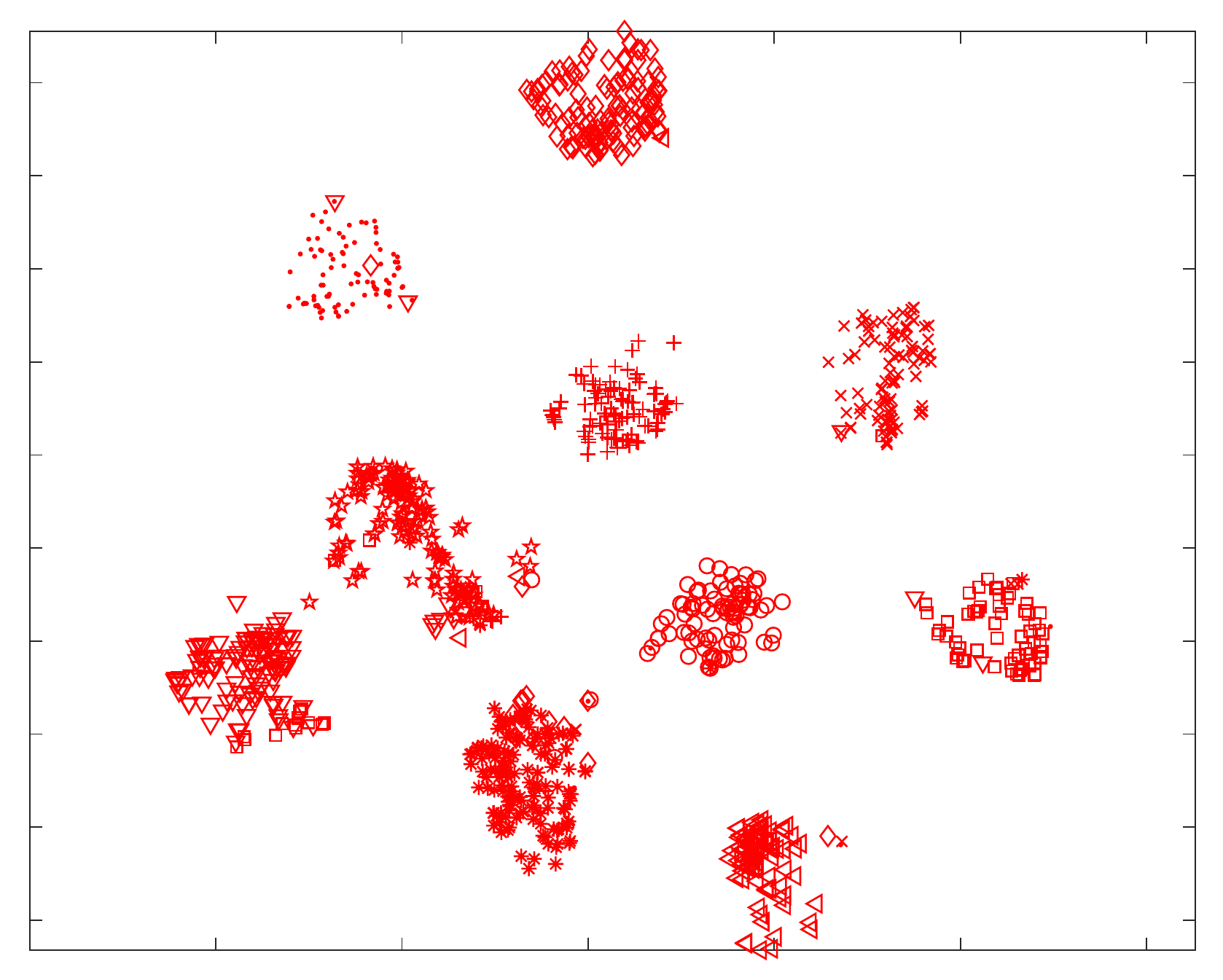}}
\end{minipage}}
\subfigure[The pseudo target domain $AC$]{\label{TSNE3-2}
\begin{minipage}[b]{0.23\textwidth}
\centering \scalebox{0.25}{ 
\includegraphics{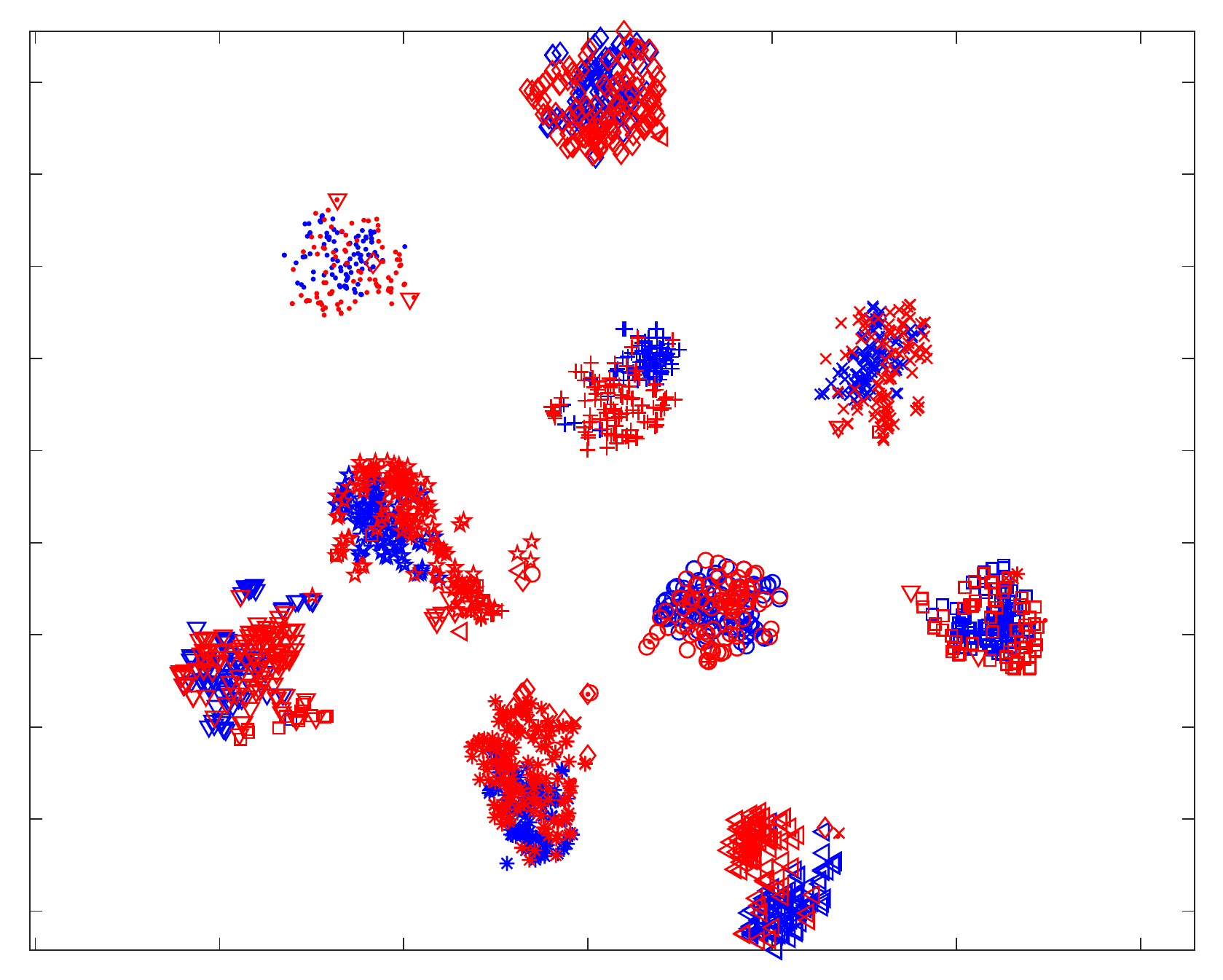}}
\end{minipage}}
\\
\subfigure[The pseudo target domain $DC$]{\label{TSNE3-1}
\begin{minipage}[b]{0.23\textwidth}
\centering \scalebox{0.25}{ 
\includegraphics{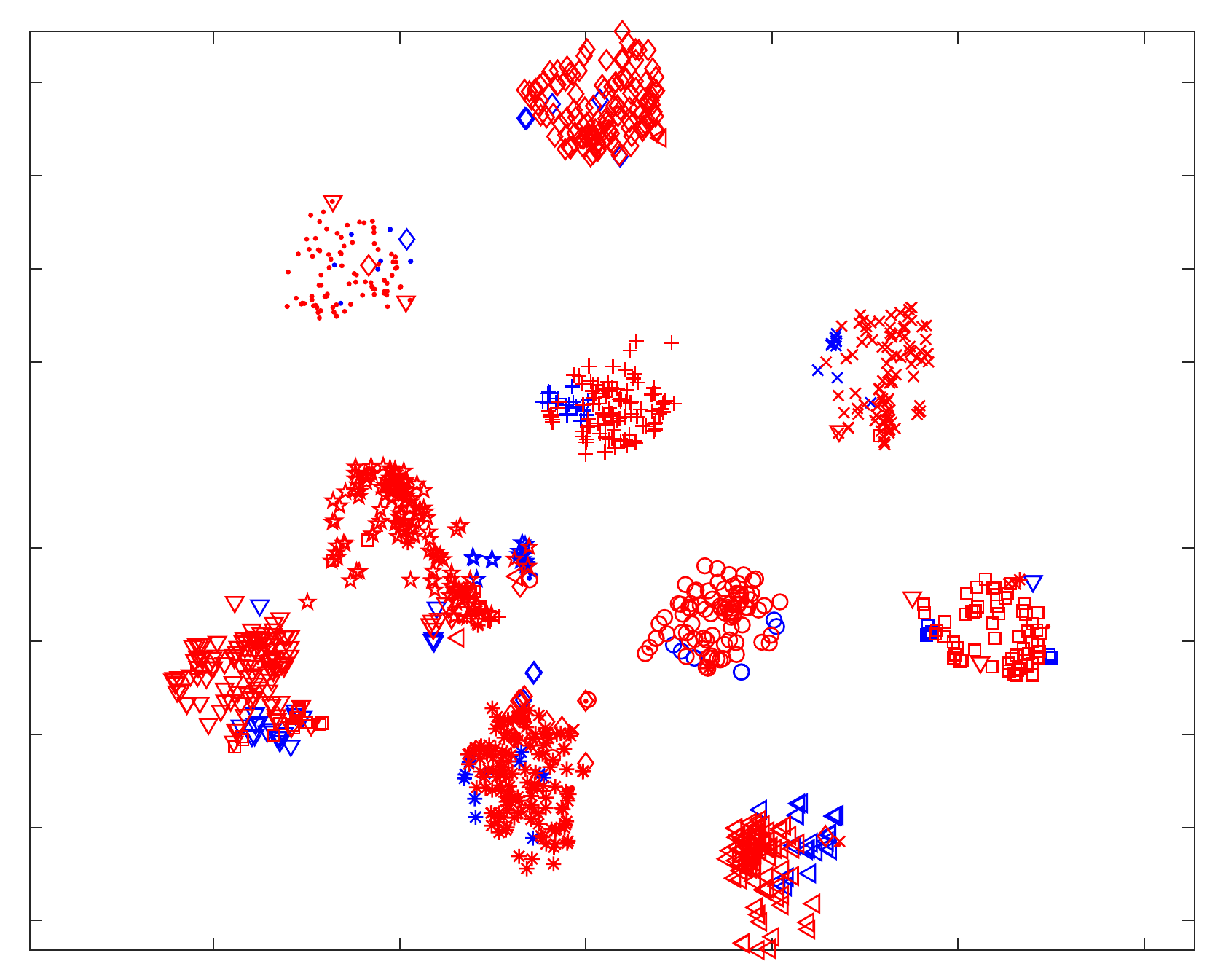}}
\end{minipage}}
\subfigure[The pseudo target domain $WC$]{\label{TSNE3-2}
\begin{minipage}[b]{0.23\textwidth}
\centering \scalebox{0.25}{ 
\includegraphics{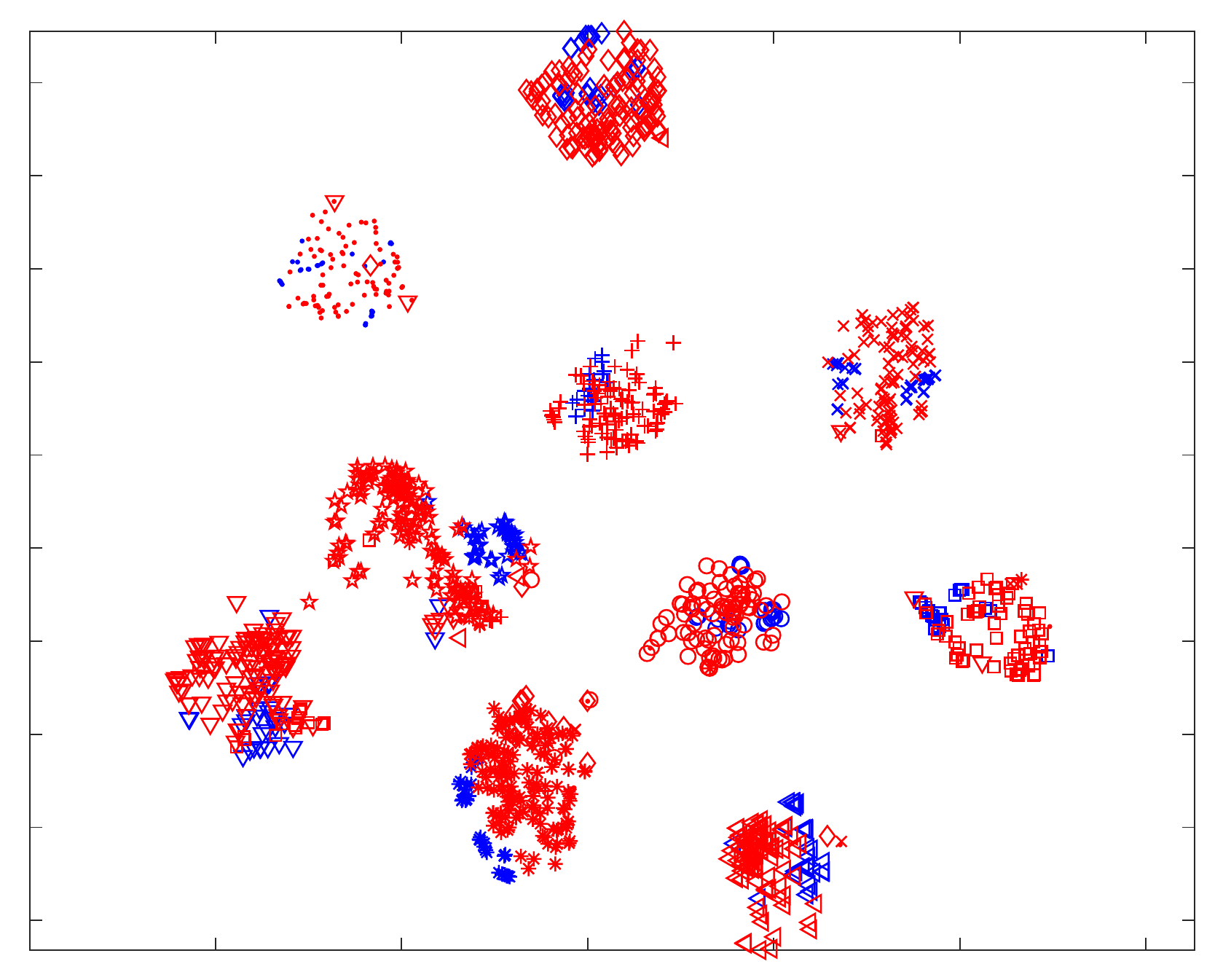}}
\end{minipage}}
\caption{Visualization on task $A,D,W \rightarrow C$ for both the actual target domain and the pseudo target domains. The actual target domain $C$ is colored in red, and other domains in the pseudo target domains are colored in blue. Features corresponding to the same category are annotated by the same marker.}\label{tsne3}}
\end{figure}

Table \ref{table5} shows the experimental results of MDA on the Digits-Five benchmark. For fair comparison, all experiments are performed on the same feature extractor architecture. ADAGE\cite{hallu2019} slightly  outperforms our PTMDA in overall average accuracy. The reason may be that ADAGE uses an elaborate Hallucinator block to remove the domain-specific style of the input images and achieves better adaptation performance. Nevertheless, our PTMDA obtains the best accuracy when using the $mt$ as the target domain. Compared with the LtC-MSDA method~\cite{Lcombine2020}, our PTMDA still achieves comparable performance. PTMDA also significantly outperforms other recent MDA methods (e.g., CMSS\cite{cmss2020}, SHOT\cite{shot2020} and MDDA\cite{DistillMSDA2020AAAI}) on most of the adaptation tasks, which confirms the improvement when using pseudo target domain for MDA task. Specifically, we achieved $6.64\%$, $0.52\%$, $3.71\%$, $3.17\%$, and $3.73\%$ accuracy improvements with respect to MDDA\cite{DistillMSDA2020AAAI} on each task, respectively. Notice that even though DCTN and M3SDA use well-designed loss term in the multi-source scenario, their performance is inferior to our PTMDA. This shows that PTMDA can sufficiently extract discriminative features for target and get a more generalized classifier, which is empirically superior for this task. Whereas the performance of DCTN is inferior to those of MCD\cite{Saito2018CVPRmcd} and MEDA\cite{Jindong2018meda}, it reveals that only reducing shifts among various domains by adversarial learning may lead to suboptimal results due to equilibrium challenge issue. Compared with DANN\cite{Yaroslav2016JMLR} in the Source-combine setting, even the most difficult task $mt,up,sv,sy \rightarrow mm$, PTMDA yields a significant improvement of $14.43\%$, which demonstrates that PTMDA can improve the performance of MDA by attenuating domain shifts cross multiple domains.

\subsection{Feature Visualization}

To visualize the PTMDA features before and after adaptation, as well as the features obtained by M3SDA, we conduct experiments by using t-SNE\cite{Maaten2008Visualizing} on task $A,D,W\rightarrow C$ on Office-Caltech10. Domains are presented in different colors for clarity. As we can see from Fig. \ref{fig7}, compared with the features of the Source-only, all the features of PTMDA show good adaptation patterns. It reveals that PTMDA can successfully learn transferable features from multiple source domains. Besides, the target features in red are easier to be classified than others. It shows that the target features learned by PTMDA achieve desirable discrimination ability. Compared with the features of M3SDA, categories in PTMDA are aligned better, and different domains show lower discrepancy. This leads to a more discriminative target feature space.

We also show the learned features for both the actual target domain $C$ and the pseudo target domains (i.e., $AC$, $DC$, and $WC$) in Fig. \ref{tsne3}. Features corresponding to the same category are visualized with the same marker. As we can see, the actual target features do separate clearly. Features of other domains (i.e., $A$, $D$, and $W$) in the pseudo target domains also show class-based discrimination, and they cluster around the corresponding actual target features. It verifies that the domain difference between the source domain and the actual target domain, inside each pseudo target domain, has been minimized actually.

\subsection{Ablation study}

In order to find that which component of PTMDA plays an important role in learning domain adaptation features, we perform ablation study over various combinations on the Digits-Five adaptation task.

We use DCTN~\cite{Xu_2018_CVPR}, M3SDA~\cite{Peng_2019_ICCV}, and MDDA~\cite{DistillMSDA2020AAAI} as the baseline methods. The method denoted by PT+BN implements the combination of pseudo target domain with the BN layers, but excludes the MN layers and the MC loss, where ``PT'' is short for ``Pseudo Target''. The method denoted by PT+MN implements the combination of pseudo target domain with the MN layers. The method denoted by PT+MC implements the combination of pseudo target domain with the MC loss and the BN layers, but excludes the MN layers. Table~\ref{table9} shows the results of ablation study.

\begin{table*}[htbp]
  \small
  \caption{Ablation study of the PTMDA method on Digits-Five}
  \centering
  \renewcommand{\tabcolsep}{0.3pc} 
  \renewcommand{\arraystretch}{1.1} 
    \begin{tabular}{ccccccc}
    \hline
   Approach &mt,up,sv,sy$\rightarrow$mm& mm,up,sv,sy$\rightarrow$mt & mm,mt,sv,sy$\rightarrow$up &  mm,mt,up,sy$\rightarrow$sv &mm,mt,up,sv$\rightarrow$sy & Avg  \\
    \hline
    DCTN \cite{Xu_2018_CVPR}  & 70.5  & 96.2  & 92.8  & 77.6  & 86.8  & 84.8  \\
    \hline
    M3SDA \cite{Peng_2019_ICCV}  & 72.8 & 98.4 & 96.1 & 81.3 & 89.6 & 87.7 \\
    \hline
    MDDA \cite{DistillMSDA2020AAAI}  & 78.6  & 98.8  & 93.9  & 79.3  & 89.7  & 88.1  \\
    \hline
    MN+MC-PT & 77.8$\pm$0.2  & 97.8$\pm$0.1  & 96.4$\pm$0.1  & 78.1$\pm$0.2  & 88.0$\pm$0.4  & 87.6$\pm$0.1  \\
    \hline
    PT+BN & 80.0$\pm$0.2  & 99.0$\pm$0.2  & 97.5$\pm$0.1  & 81.6$\pm$0.4  & 91.6$\pm$0.3  & 89.9$\pm$0.1  \\
    \hline
    PT+MN & 82.8$\pm$0.3  & 99.0$\pm$0.2  & 97.8$\pm$0.3  & 82.0$\pm$0.3  & 92.1$\pm$0.3  & 90.7$\pm$0.1  \\
    \hline
    PT+MC & 84.3$\pm$0.2  & 99.1$\pm$0.3  & 97.4$\pm$0.3  & 82.0$\pm$0.2  & 93.4$\pm$0.3  & 91.2$\pm$0.1  \\
    \hline
    \textbf{PTMDA(PT+MN+MC) } & \textbf{85.2$\pm$0.4}  & \textbf{99.3$\pm$0.1}  & \textbf{97.6$\pm$0.3}  & \textbf{82.5$\pm$0.5} & \textbf{93.4$\pm$0.3} & \textbf{91.6$\pm$0.2}  \\
    \hline
    \end{tabular}%
  \label{table9}%
\end{table*}

Comparing with DCTN~\cite{Xu_2018_CVPR}, M3SDA~\cite{Peng_2019_ICCV}, and MDDA~\cite{DistillMSDA2020AAAI}, PT+BN performs better in the average sense, which demonstrates the effectiveness of pseudo target domain. Notice that PT+MN obtains an improvement of 0.8\% over PT+BN. It indicates that the distribution alignments between the whitened source domains and the target domain enhances the generalization ability of DNNs. PT+MC achieves 6.4\%, 3.1\%, and 1.3\% improvements against DCTN, MDDA, and PT+BN, respectively. Since they are adversarial based methods, these results verify that the MC loss can moderate the equilibrium challenge and improve the performance of adversarial learning. PT+MC has a larger improvement than PT+MN, which indicates that the MC loss makes greater contribution to the MDA tasks. Comparing PT+MC with PTMDA, the average accuracy is improved from 91.2\% to 91.6\%. The result validates the effectiveness of our proposed MN layers. From the results shown above, we confirm that each component in PTMDA has its specific contribution. By combining pseudo target domain with MN layers and the MC loss, PTMDA leads to further improvement. All these results indicate their complementarity and superiority of the PTMDA method.

We also provide the experimental results obtained by only using the real target domain with pseudo labels (i.e., no source domain in the pseudo target domain) in Table~\ref{table9}, and it is denoted as MN+MC-PT. We can see that PTMDA outperforms MN+MC-PT by a large margin (91.6\% vs. 87.6\%). It means that the pseudo target domain plays an important role in improving the performance of the MDA tasks.

\subsection{Other analysis}

We conduct experiments with different numbers (i.e., two or three) of source domains on two MDA tasks on Office-Caltech10. The results are shown in Table~\ref{Tab1number}. In terms of the overall performance, the average accuracy which is obtained by using three source domains is slightly better than those of using two source domains. It seems that the number of source domains has little influence on the performance of PTMDA.

\begin{table*}[htbp]
  \small
  \caption{Accuracy (\%) comparison among different numbers and different orders of source domains for $A,D,W \rightarrow C$ and $C,D,W \rightarrow A$ on Office-Caltech10.}
  \centering
  \renewcommand{\tabcolsep}{0.4pc} 
  \renewcommand{\arraystretch}{1.1} 
    \begin{tabular}{ccc|cccccc}
    \hline
     A,D,W $\rightarrow$ C & D,A,W $\rightarrow$ C & W,A,D $\rightarrow$ C& A,W $\rightarrow$ C & W,A $\rightarrow$ C & A,D $\rightarrow$ C& D,A $\rightarrow$ C & W,D $\rightarrow$ C & D,W $\rightarrow$ C \\
    \hline
     96.5$\pm$0.2 & 96.5$\pm$0.2& 96.5$\pm$0.2 & 96.1$\pm$0.1& 96.1$\pm$0.1 &95.8$\pm$0.1& 95.8$\pm$0.1 & 94.9$\pm$0.1 & 94.9$\pm$0.1\\
    \hline
     C,D,W $\rightarrow$ A &D,C,W $\rightarrow$ A &W,D,C $\rightarrow$ A & C,W $\rightarrow$ A & W,C $\rightarrow$ A & D,C $\rightarrow$ A& C,D $\rightarrow$ A & W,D $\rightarrow$ A & D,W $\rightarrow$ A \\
    \hline
     96.7$\pm$0.4 &96.7$\pm$0.4&96.7$\pm$0.4 & 96.3$\pm$0.1& 96.3$\pm$0.1 &96.2$\pm$0.1 &96.2$\pm$0.1 & 95.8$\pm$0.2 & 95.8$\pm$0.2\\
    \hline
    \end{tabular}%
  \label{Tab1number}%
\end{table*}

To study the impact of different orders of source domains, we conduct additional experiments. For the Office-Caltech10 dataset, we use different orders of source domains for $A,D,W \rightarrow C$ and $C,D,W \rightarrow A$ with different numbers (i.e., two or three) of source domains, and the results are shown in Table~\ref{Tab1number}. For the Digits-five dataset, among the twenty-four permutations of four source domains, we just randomly choose six permutations for the task $mt,up,sv,sy \rightarrow mm$ due to the time afford. The results are shown in Table~\ref{table18orders}. We run each experiment five times, and find that different orders of source domains of each task get identical average accuracy. It is consistent with our intuition. Actually, each group of the source domain and target domain is used in turn to construct a pseudo target domain, and samples are randomly sampled from each source domain. Thus, the order of the source-target adaption task does not affect the final performance.

\begin{table}[htbp]
  \small
  \caption{Accuracy (\%) comparison among different orders of source domains for $mt,\!up,\!sv,\!sy\!\!\rightarrow\!\!mm$ on Digits-Five}
  \centering
  \renewcommand{\tabcolsep}{0.3pc} 
  \renewcommand{\arraystretch}{1.1} 
    \begin{tabular}{cccc}
    \hline
     mt,up,sv,sy$\rightarrow$mm & mt,sv,up,sy$\rightarrow$mm & up,sy,sv,mt$\rightarrow$mm \\
    \hline
     85.2$\pm$0.4 &85.2$\pm$0.3 & 85.2$\pm$0.2 \\
    \hline
     sv,up,sy,mt$\rightarrow$mm & sy,mt,sv,up$\rightarrow$mm & up,sv,sy,mt$\rightarrow$mm \\
    \hline
     85.2$\pm$0.3 &85.2$\pm$0.3 & 85.2$\pm$0.2 \\
    \hline
    \end{tabular}%
  \label{table18orders}%
\end{table}

\section{Conclusion}\label{section5}

In this paper, we propose Pseudo Target for MDA (PTMDA), in which we construct a pseudo target domain to mimic a new domain in a group-specific subspace and align the remainder source domains with the pseudo target domain. PTMDA can sufficiently extract structural and relevant information among multiple sources to promote transfer efficiency, and improve the performance of classifier on the real target domain. To further enhance the transferability of deep neural networks, we design a matching normalization layer to align the feature distributions of different domains in the intermediate layers of the feature extractor. Extensive experiments on several benchmarks validate that PTMDA can outperform or compete state-of-the-art methods. Ablation study shows that each component in PTMDA has specific contribution to the MDA task. In the future, we plan to extend this method to the scenario with label shift.

\bibliographystyle{IEEEtran}
\bibliography{ref-PTMDA}

\begin{thebibliography}{10}
\providecommand{\url}[1]{#1}
\csname url@samestyle\endcsname
\providecommand{\newblock}{\relax}
\providecommand{\bibinfo}[2]{#2}
\providecommand{\BIBentrySTDinterwordspacing}{\spaceskip=0pt\relax}
\providecommand{\BIBentryALTinterwordstretchfactor}{4}
\providecommand{\BIBentryALTinterwordspacing}{\spaceskip=\fontdimen2\font plus
\BIBentryALTinterwordstretchfactor\fontdimen3\font minus
  \fontdimen4\font\relax}
\providecommand{\BIBforeignlanguage}[2]{{%
\expandafter\ifx\csname l@#1\endcsname\relax
\typeout{** WARNING: IEEEtran.bst: No hyphenation pattern has been}%
\typeout{** loaded for the language `#1'. Using the pattern for}%
\typeout{** the default language instead.}%
\else
\language=\csname l@#1\endcsname
\fi
#2}}
\providecommand{\BIBdecl}{\relax}
\BIBdecl

\bibitem{resnet2016}
K.~M. He, X.~Y. Zhang, S.~Q. Ren \emph{et~al.}, ``Deep residual learning for
  image recognition,'' in \emph{CVPR}, 2016, pp. 770--778.

\bibitem{zhanglei2020Class}
S.~S. Wang, L.~Zhang, W.~M. Zuo \emph{et~al.}, ``Class-specific reconstruction
  transfer learning for visual recognition across domains,'' \emph{IEEE Trans.
  Image Process.}, vol.~29, pp. 2424--2438, 2020.

\bibitem{Yang2020tip}
S.~T. Chen, M.~Harandi, X.~N. Jin \emph{et~al.}, ``Domain adaptation by joint
  distribution invariant projections,'' \emph{IEEE Trans. Image Process.},
  vol.~29, pp. 8264--8277, 2020.

\bibitem{ren2014tipTransfer}
C.-X. Ren, D.-Q. Dai, K.-K. Huang \emph{et~al.}, ``Transfer learning of
  structured representation for face recognition,'' \emph{IEEE Trans. Image
  Process.}, vol.~23, no.~12, pp. 5440--5454, 2014.

\bibitem{Lu2016tip}
J.~L. Hu, J.~W. Lu, Y.~P. Tan \emph{et~al.}, ``Deep transfer metric learning,''
  \emph{IEEE Trans. Image Process.}, vol.~25, no.~12, pp. 5576--5588, 2016.

\bibitem{Yaroslav2016JMLR}
Y.~Ganin, E.~Ustinova, H.~Ajakan \emph{et~al.}, ``Domain-adversarial training
  of neural networks,'' \emph{JMLR}, vol.~17, no.~59, pp. 1--35, 2016.

\bibitem{Saito2018CVPRmcd}
K.~Saito, K.~Watanabe, Y.~Ushiku \emph{et~al.}, ``Maximum classifier
  discrepancy for unsupervised domain adaptation,'' in \emph{CVPR}, 2018, pp.
  3723--3732.

\bibitem{ren2020TCYB1}
C.-X. Ren, X.-L. Xu, and H.~Yan, ``Generalized conditional domain adaptation: A
  causal perspective with low-rank translators,'' \emph{IEEE Trans. Cyber.},
  vol.~50, no.~2, pp. 821--834, 2020.

\bibitem{Kui2019CVPR}
Y.~B. Zhang, H.~Tang, K.~Jia \emph{et~al.}, ``Domain-symmetric networks for
  adversarial domain adaptation,'' in \emph{CVPR}, 2019, pp. 5026--5035.

\bibitem{ren2019TCYB1}
C.-X. Ren, J.~S. Feng, D.-Q. Dai, and S.~Yan, ``Heterogenous domain adaptation
  via covariance structured feature translators,'' \emph{IEEE Trans. Cyber.},
  vol.~51, no.~4, pp. 2166--2177, 2021.

\bibitem{Long2019PAMI1}
M.~S. Long, Y.~Cao, Z.~J. Cao \emph{et~al.}, ``Transferable representation
  learning with deep adaptation networks,'' \emph{IEEE Trans. Pattern Anal. and
  Mach. Intell.}, vol.~41, no.~12, pp. 3071--3085, 2019.

\bibitem{mdd2020pami1}
J.~J. Li, E.~P. Chen, Z.~M. Ding \emph{et~al.}, ``Maximum density divergence
  for domain adaptation,'' \emph{IEEE Trans. Pattern Anal. and Mach. Intell.},
  vol.~43, no.~11, pp. 3918--3930, 2021.

\bibitem{msda2012video1}
L.~X. Duan, D.~Xu, and I.~W.-H. Tsang, ``Domain adaptation from multiple
  sources: A domain-dependent regularization approach,'' \emph{IEEE Trans.
  Neural Netw. and Learning Syst.}, vol.~23, no.~3, pp. 504--518, 2012.

\bibitem{zhao2019NIPS1seg}
S.~C. Zhao, B.~Li, X.~Y. Yue \emph{et~al.}, ``Multi-source domain adaptation
  for semantic segmentation,'' in \emph{NeurIPS}, 2019, pp. 7285--7298.

\bibitem{Mansour2009NIPS1}
Y.~Mansour, M.~Mohri, and A.~Rostamizadeh, ``Domain adaptation with multiple
  sources,'' in \emph{NeurIPS}, 2009, pp. 1041--1048.

\bibitem{Xu_2018_CVPR}
R.~J. Xu, Z.~L. Chen, W.~M. Zuo \emph{et~al.}, ``Deep cocktail network:
  Multi-source unsupervised domain adaptation with category shift,'' in
  \emph{CVPR}, 2018, pp. 3964--3973.

\bibitem{zhu2019MSDA}
Y.~C. Zhu, F.~Z. Zhuang, and D.~Q. Wang, ``Aligning domain-specific
  distribution and classifier for cross-domain classification from multiple
  sources,'' in \emph{AAAI}, 2019.

\bibitem{DistillMSDA2020AAAI}
S.~C. Zhao, G.~Z. Wang, S.~H. Zhang \emph{et~al.}, ``Multi-source distilling
  domain adaptation,'' in \emph{AAAI}, 2020.

\bibitem{Zhao2018NIPS}
H.~Zhao, S.~H. Zhang, G.~H. Wu \emph{et~al.}, ``Adversarial multiple source
  domain adaptation,'' in \emph{NeurIPS}, 2018, pp. 8559--8570.

\bibitem{Hoffman2018NIPS}
J.~Hoffman, M.~Mohri, and N.~S. Zhang, ``Algorithms and theory for
  multiple-source adaptation,'' in \emph{NeurIPS}, 2018, pp. 8246--8256.

\bibitem{Ievgen2019On}
I.~Redko, A.~Habrard, and M.~Sebban, ``On the analysis of adaptability in
  multi-source domain adaptation,'' \emph{Mach. Learning}, no.~2, pp.
  1635--1652, 2019.

\bibitem{Hoffman2012LtnDm}
J.~Hoffman, B.~Kulis, T.~Darrell \emph{et~al.}, ``Discovering latent domains
  for multisource domain adaptation,'' in \emph{ECCV}, 2012, pp. 702--715.

\bibitem{Peng_2019_ICCV}
X.~C. Peng, Q.~X. Bai, X.~D. Xia \emph{et~al.}, ``Moment matching for
  multi-source domain adaptation,'' in \emph{ICCV}, 2019, pp. 1406--1415.

\bibitem{chen2020Multiple}
C.~Q. Chen, W.~P. Xie, Y.~Wen \emph{et~al.}, ``Multiple-source domain
  adaptation with generative adversarial nets,'' \emph{Knowledge-Based
  Systems}, vol. 199, p. 105962, 2020.

\bibitem{BoostMda2018}
M.~Mancini, L.~Porzi, S.~R. Bulo \emph{et~al.}, ``Boosting domain adaptation by
  discovering latent domains,'' in \emph{CVPR}, 2018, pp. 3771--3780.

\bibitem{DSBN2019}
W.-G. Chang, T.~You, S.~Seo \emph{et~al.}, ``Domain-specific batch
  normalization for unsupervised domain adaptation,'' in \emph{CVPR}, 2019, pp.
  7346--7354.

\bibitem{MultiDial2020}
F.~M. Carlucci, L.~Porzi, B.~Caputo \emph{et~al.}, ``Multidial: Domain
  alignment layers for (multisource) unsupervised domain adaptation,''
  \emph{IEEE Trans. Pattern Anal. and Mach. Intell.}, vol.~43, no.~12, pp.
  4441--4452, 2021.

\bibitem{2015Batch}
S.~Ioffe and C.~Szegedy, ``Batch normalization: Accelerating deep network
  training by reducing internal covariate shift,'' in \emph{ICML}, 2015, pp.
  448--456.

\bibitem{Ben2006NIPS}
S.~Ben-David, J.~Blitzer, K.~Crammer \emph{et~al.}, ``Analysis of
  representations for domain adaptation,'' in \emph{NeurIPS}, 2006, pp.
  137--144.

\bibitem{Blitzer2007NIPS}
J.~Blitzer, K.~Crammer, A.~Kulesza \emph{et~al.}, ``Learning bounds for domain
  adaptation,'' in \emph{NeurIPS}, 2008, pp. 129--136.

\bibitem{uniDA2020}
K.~Saito, D.~Kim, S.~Sclaroff \emph{et~al.}, ``Universal domain adaptation
  through self supervision,'' in \emph{NeurIPS}, 2020, pp. 16\,282--16\,292.

\bibitem{Li2020etd}
M.~X. Li, Y.~M. Zhai, Y.~W. Luo \emph{et~al.}, ``Enhanced transport distance
  for unsupervised domain adaptation,'' in \emph{CVPR}, 2020, pp.
  13\,933--13\,941.

\bibitem{Kui2018IJCAI}
Y.~G. Yan, W.~Li, H.~R. Wu \emph{et~al.}, ``Semi-supervised optimal transport
  for heterogeneous domain adaptation,'' in \emph{Intern. Joint Conf. on
  Artificial Intell.}, 2018, pp. 2969--2975.

\bibitem{YouWeiPami1}
Y.-W. Luo, C.-X. Ren, D.-Q. Dai, and Y.~Hong, ``Unsupervised domain adaptation
  via discriminative manifold propagation,'' \emph{IEEE Trans. Pattern Anal.
  and Mach. Intell.}, vol.~44, no.~3, pp. 1653--1669, 2022.

\bibitem{TgtInd2020}
P.~Pandey, P.~A. P, V.~Kyatham \emph{et~al.}, ``Target-independent domain
  adaptation for wbc classification using generative latent search,''
  \emph{IEEE Trans. Medical Imag.}, vol.~39, no.~12, pp. 3979--3991, 2020.

\bibitem{Tzeng2017CVPRadda}
E.~Tzeng, J.~Hoffman, K.~Saenko \emph{et~al.}, ``Adversarial discriminative
  domain adaptation,'' in \emph{CVPR}, 2017, pp. 7167--7176.

\bibitem{hallu2019}
F.~M. Carlucci, P.~Russo, T.~Tommasi \emph{et~al.}, ``Hallucinating agnostic
  images to generalize across domains,'' in \emph{ICCV Workshop}, 2019, pp.
  3227--3234.

\bibitem{lookback2019}
V.~K. Kurmi and V.~P. Namboodiri, ``Looking back at labels: A class based
  domain adaptation technique,'' in \emph{Intern. Joint Conf. on Neural Netw.
  (IJCNN)}, 2019, pp. 1--8.

\bibitem{pei2018multi}
Z.~Pei, Z.~Cao, M.~Long \emph{et~al.}, ``Multi-adversarial domain adaptation,''
  in \emph{AAAI}, 2018.

\bibitem{Asymm2017}
K.~Saito, Y.~Ushiku, and T.~Harada, ``Asymmetric tri-training for unsupervised
  domain adaptation,'' in \emph{ICML}, 2017, pp. 2988--2997.

\bibitem{Collab2021}
W.~Zhang, D.~Xu, W.~Ouyang \emph{et~al.}, ``Self-paced collaborative and
  adversarial network for unsupervised domain adaptation,'' \emph{IEEE Trans.
  Pattern Anal. and Mach. Intell.}, vol.~43, no.~6, pp. 2047--2061, 2021.

\bibitem{Collab2018}
W.~Zhang, W.~Ouyang, W.~Li \emph{et~al.}, ``Collaborative and adversarial
  network for unsupervised domain adaptation,'' in \emph{CVPR}, 2018, pp.
  3801--3809.

\bibitem{Trip2021}
W.~Deng, L.~Zheng, Y.~Sun \emph{et~al.}, ``Rethinking triplet loss for domain
  adaptation,'' \emph{IEEE Trans. Circ. and Syst. for Video Tech.}, vol.~31,
  no.~1, pp. 29--37, 2021.

\bibitem{AggrMSDA2020ICLM}
J.~F. Wen, R.~Greiner, and D.~Schuurmans, ``Domain aggregation networks for
  multi-source domain adaptation,'' in \emph{ICML}, 2020, pp. 10\,214--10\,224.

\bibitem{cdan1}
M.~S. Long, Z.~J. Cao, J.~M. Wang \emph{et~al.}, ``Conditional adversarial
  domain adaptation,'' in \emph{NeurIPS}, 2018, pp. 1640--1650.

\bibitem{2017gan}
S.~Arora, R.~Ge, Y.~Y. Liang \emph{et~al.}, ``Generalization and equilibrium in
  generative adversarial nets ({GAN}s),'' in \emph{ICML}, 2017, pp. 224--232.

\bibitem{Fei2019DUAE}
P.~F. Ge, C.~X. Ren, D.~Q. Dai \emph{et~al.}, ``Dual adversarial autoencoders
  for clustering,'' \emph{IEEE Trans. Neural Netw. and Learn. Syst.}, vol.~31,
  no.~4, pp. 1417--1424, 2020.

\bibitem{GenUD2021}
P.~Pandey, M.~Raman, S.~Varambally \emph{et~al.}, ``Generalization on unseen
  domains via inference-time label-preserving target projections,'' in
  \emph{CVPR}, 2021, pp. 12\,919--12\,928.

\bibitem{kim2021selfreg}
D.~Kim, Y.~Yoo, S.~Park \emph{et~al.}, ``Selfreg: Self-supervised contrastive
  regularization for domain generalization,'' in \emph{ICCV}, 2021, pp.
  9619--9628.

\bibitem{RenDos2015}
C.-X. Ren, D.-Q. Dai, X.~He \emph{et~al.}, ``Sample weighting: An inherent
  approach for outlier suppressing discriminant analysis,'' \emph{IEEE Trans.
  Knowl. and Data Engineering}, vol.~27, no.~11, pp. 3070--3083, 2015.

\bibitem{2018How}
S.~Santurkar, D.~Tsipras, A.~Ilyas \emph{et~al.}, ``How does batch
  normalization help optimization?'' in \emph{NeurIPS}, 2018, pp. 2483--2493.

\bibitem{AdaBN2018pr}
Y.~H. Li, N.~Y. Wang, J.~P. Shi \emph{et~al.}, ``Adaptive batch normalization
  for practical domain adaptation,'' \emph{Pattern Recog.}, vol.~80, pp.
  109--117, 2018.

\bibitem{miniSGD2014}
M.~Li, T.~Zhang, Y.~Q. Chen \emph{et~al.}, ``Efficient mini-batch training for
  stochastic optimization,'' in \emph{The 20th ACM SIGKDD}, 2014, pp. 661--670.

\bibitem{xavier2010}
X.~Glorot and Y.~Bengio, ``Understanding the difficulty of training deep
  feedforward neural networks,'' in \emph{Intern. Conf. on Artificial Intell.
  and Stat.}, vol.~9, 2010, pp. 249--256.

\bibitem{Ben2010MacLen}
S.~Ben-David, J.~Blitzer, K.~Crammer \emph{et~al.}, ``A theory of learning from
  different domains,'' \emph{Mach. Learning}, vol.~79, no.~1, pp. 151--175,
  2010.

\bibitem{DANN2015}
Y.~Ganin and V.~Lempitsky, ``Unsupervised domain adaptation by
  backpropagation,'' in \emph{ICML}, 2015, pp. 1180--1189.

\bibitem{Bsp2019}
X.~Chen, S.~Wang, M.~Long \emph{et~al.}, ``Transferability vs.
  discriminability: Batch spectral penalization for adversarial domain
  adaptation,'' in \emph{ICML}, 2019, pp. 1081--1090.

\bibitem{Saenko2010da1}
K.~Saenko, B.~Kulis, M.~Fritz \emph{et~al.}, ``Adapting visual category models
  to new domains,'' in \emph{ECCV}, 2010, pp. 213--226.

\bibitem{DSAN2020}
Y.~Zhu, F.~Zhuang, J.~Wang \emph{et~al.}, ``Deep subdomain adaptation network
  for image classification,'' \emph{IEEE Trans. Neural Netw. and Learn. Syst.},
  vol.~32, no.~4, pp. 1713--1722, 2020.

\bibitem{coral2016AAAI}
B.~C. Sun, J.~S. Feng, and K.~Saenko, ``Return of frustratingly easy domain
  adaptation,'' in \emph{AAAI}, 2016.

\bibitem{LongM2017jda}
M.~S. Long, H.~Zhu, J.~M. Wang \emph{et~al.}, ``Deep transfer learning with
  joint adaptation networks,'' in \emph{ICML}, 2017, pp. 2208--2217.

\bibitem{Jindong2018meda}
J.~D. Wang, W.~J. Feng, Y.~Q. Chen \emph{et~al.}, ``Visual domain adaptation
  with manifold embedded distribution alignment,'' in \emph{The 26th ACM
  Intern. Conf. on Multimedia}, 2018, pp. 402--410.

\bibitem{2021MADAN}
S.~Zhao, B.~Li, P.~Xu \emph{et~al.}, ``Madan: Multi-source adversarial domain
  aggregation network for domain adaptation,'' \emph{ICCV}, vol. 129, pp.
  1--26, 2021.

\bibitem{German1}
S.~Rakshit, B.~Banerjee, G.~Roig \emph{et~al.}, ``Unsupervised multi-source
  domain adaptation driven by deep adversarial ensemble learning,'' in
  \emph{The German Conf. on Pattern Recog.}, 2019, pp. 485--498.

\bibitem{Lcombine2020}
H.~Wang, M.~Xu, B.~Ni \emph{et~al.}, ``Learning to combine: Knowledge
  aggregation for multi-source domain adaptation,'' in \emph{ECCV}, 2020, pp.
  727--744.

\bibitem{cmss2020}
L.~Yang, Y.~Balaji, S.-N. Lim \emph{et~al.}, ``Curriculum manager for source
  selection in multi-source domain adaptation,'' in \emph{ECCV}, 2020, pp.
  608--624.

\bibitem{shot2020}
J.~Liang, D.~Hu, and J.~Feng, ``Do we really need to access the source data?
  {S}ource hypothesis transfer for unsupervised domain adaptation,'' in
  \emph{ICML}, 2020, pp. 6028--6039.

\bibitem{Maaten2008Visualizing}
L.~V.~D. Maaten and G.~E. Hinton, ``Visualizing data using t-{SNE},''
  \emph{JMLR}, vol.~9, pp. 2579--2605, 2008.

\end{thebibliography}

\end{document}